\documentclass{article}
\PassOptionsToPackage{numbers,sort&compress}{natbib}
\usepackage[final]{COLM_template/colm2026_conference}
\setcitestyle{numbers,square,comma}

\usepackage{microtype}
\usepackage{hyperref}
\usepackage{lineno}

\definecolor{darkblue}{rgb}{0, 0, 0.5}
\hypersetup{colorlinks=true, citecolor=darkblue, linkcolor=darkblue, urlcolor=darkblue}

\usepackage[svgnames]{xcolor}
\usepackage[table]{xcolor}
\definecolor{DarkGreen}{rgb}{0.2,0.63,0.17}
\definecolor{DarkRed}{rgb}{0.89,0.1,0.11}
\definecolor{DarkBlue}{rgb}{0.12,0.47,0.71}
\definecolor{Gray}{rgb}{0.2,0.2,0.2}
\definecolor{bestcolor}{RGB}{214,232,213}
\definecolor{secondcolor}{RGB}{173,216,230}
\definecolor{rqonecolor}{RGB}{6,120,29}
\definecolor{rqtwocolor}{RGB}{227,27,30}

\definecolor{DarkGreen}{RGB}{98,179,149}
\definecolor{DarkRed}{RGB}{206,86,69}
\definecolor{SABcolor}{RGB}{214,232,213}

\definecolor{bestcolor}{RGB}{214,232,213}
\definecolor{worstcolor}{RGB}{252,214,214}

\RequirePackage{amsthm,amsmath,amsfonts,amssymb}
\usepackage{url}

\RequirePackage{graphicx}

\usepackage{multicol}

\usepackage{wrapfig}

\usepackage{multirow}
\usepackage{pdflscape}

\usepackage{comment}
\usepackage{longtable}
\usepackage{adjustbox}
\usepackage[T1]{fontenc}
\usepackage[utf8]{inputenc}

\usepackage{tcolorbox}
\usepackage{minted}
\tcbuselibrary{minted,skins}

\usepackage{listings}

\definecolor{mycolor}{HTML}{1f77b4}
\tcbuselibrary{listings}
\tcbuselibrary{breakable}

\lstdefinestyle{promptboxstyle}{
  basicstyle=\ttfamily\scriptsize,
  breaklines=true,
  keywordstyle=\color{blue},
  commentstyle=\itshape\color{teal!60!black},
  stringstyle=\color{purple!70!black}
}

\newtcblisting{promptbox}[1][]{
  colback=mycolor!5!white,
  colframe=mycolor,
  listing only,
  listing options={style=promptboxstyle, frame=none},
  title=\centering \textsf{#1},
  breakable
}

\usepackage{array}

\usepackage{tikz}
\usetikzlibrary{decorations.pathreplacing}
\usepackage{subcaption}
\usepackage{enumitem}

\usepackage{fontawesome5}

\usepackage{booktabs}
\usepackage{tcolorbox}
\usepackage{arydshln}

\usepackage{pifont}
\usepackage{wasysym}
\newcommand{\cmark}{\ding{51}}%
\newcommand{\xmark}{\ding{55}}%
\newcommand{\wcircle}{{$\RIGHTcircle$}}

\newtcolorbox{practicalbox}{
  colback=blue!5,    
  colframe=cyan!99!black, 
  coltitle=black,
  title=Practical Recommendation,
  fonttitle=\bfseries,
  sharp corners,
  boxrule=0.8pt,
  before=\par\medskip\noindent,
  after=\par\medskip
}
\newtcolorbox{practicalboxnotitle}{
  colback=blue!5,    
  colframe=cyan!99!black, 
  coltitle=black,
  fonttitle=\bfseries,
  sharp corners,
  boxrule=0.8pt,
  before=\par\medskip\noindent,
  after=\par\medskip
}

\definecolor{figurebg}{RGB}{248,248,250}
\definecolor{figureborder}{RGB}{210,210,218}
\definecolor{usermsgbg}{RGB}{232,243,255}
\definecolor{agentmsgbg}{RGB}{243,243,243}
\definecolor{systemmsgbg}{RGB}{255,235,235}

\newtcolorbox{userturn}{
  colback=usermsgbg, colframe=DarkBlue!60,
  boxrule=0pt, leftrule=2pt, sharp corners,
  left=3mm, right=2mm, top=1.5mm, bottom=1.5mm,
  before skip=1.5mm, after skip=0mm,
  fontupper=\small
}

\newtcolorbox{agentturn}{
  colback=agentmsgbg, colframe=Gray,
  boxrule=0pt, leftrule=2pt, sharp corners,
  left=3mm, right=2mm, top=1.5mm, bottom=1.5mm,
  before skip=1.5mm, after skip=0mm,
  fontupper=\small
}

\newtcolorbox{systurn}{
  colback=systemmsgbg, colframe=DarkRed!60,
  boxrule=0pt, leftrule=2pt, sharp corners,
  left=3mm, right=2mm, top=1.5mm, bottom=1.5mm,
  before skip=1.5mm, after skip=0mm,
  fontupper=\small\itshape
}

\newtcolorbox{intentturn}[1]{
  enhanced,
  colback=usermsgbg, colframe=DarkBlue!60,
  boxrule=0pt, leftrule=2pt, sharp corners,
  left=3mm, right=20mm, top=1.5mm, bottom=1.5mm,
  before skip=3mm, after skip=0mm,
  fontupper=\small,
  overlay={\node[anchor=north east, inner sep=1.5pt, xshift=-1.5mm, yshift=-1.5mm]
    at (frame.north east) {\fbox{\footnotesize\strut\texttt{#1}}};}
}

\newcommand{\toolcall}[2]{%
  \\\hspace*{2mm}{\footnotesize{\color{Gray}$\triangleright$}~\texttt{#1} \textit{#2}}%
}
\newcommand{\usericon}{{\small\color{DarkBlue}\faUser}~}
\newcommand{\agenticon}{{\small\color{Gray}\faRobot}~}

\newcommand{\dataset}{\texttt{SWE-chat}}
\newcommand{\website}{\href{https://swe-chat.com}{Website}}
\newcommand{\dataurl}{https://huggingface.co/datasets/SALT-NLP/SWE-chat}
\newcommand{\codeurl}{https://github.com/SALT-NLP/SWE-chat}

\title{\dataset{}:\\Coding Agent Interactions From Real Users in the Wild}

\author{Joachim Baumann, Vishakh Padmakumar, Xiang Li, John Yang,\\
\textbf{Diyi Yang\textsuperscript{$*$} \&  Sanmi Koyejo}\thanks{Equal advising.} \\
Stanford University \\
\texttt{\{baumann,diyiy,sanmi\}@cs.stanford.edu}
\\
\\
\begin{tabular}{c}
\raisebox{-0.2\height}{\includegraphics[height=1em]{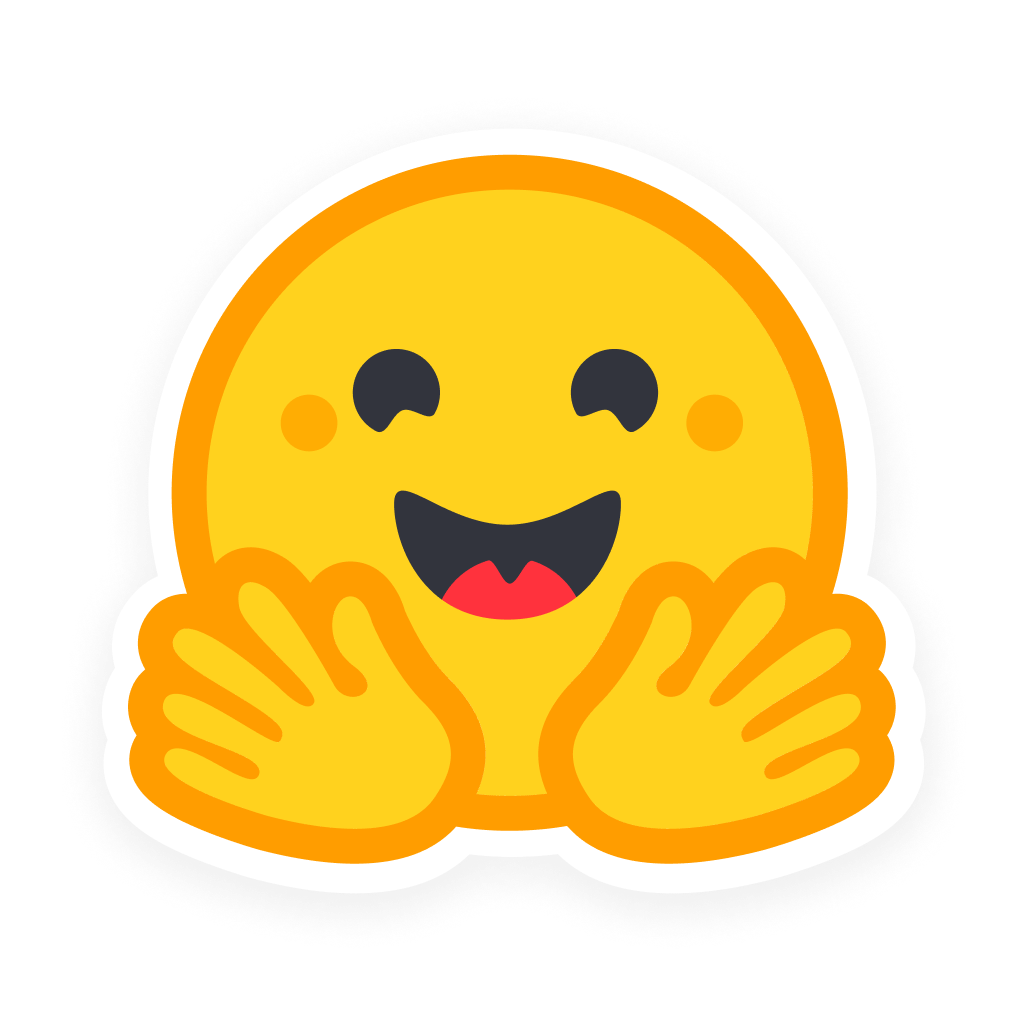}} \href{\dataurl}{Data} \hspace{0.3cm}
\raisebox{-0.2\height}{\includegraphics[height=1em]{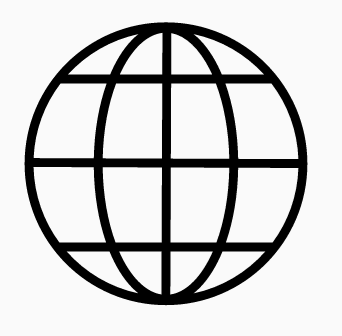}} \website \hspace{0.3cm}
\end{tabular}
\raisebox{-0.2\height}{\includegraphics[height=1em]{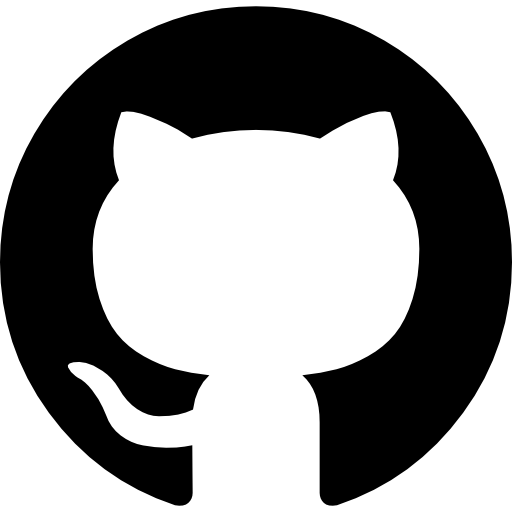}} \href{\codeurl}{Code} \hspace{0.3cm}
}

\begin{document}

\ifcolmsubmission
\linenumbers
\fi

\maketitle

\begin{abstract}
AI coding agents are being adopted at scale, yet we lack empirical evidence on how people actually use them and how much of their output is useful in practice.
We present \dataset{}, the first large-scale dataset of real coding agent sessions collected from open-source developers in the wild.
The dataset currently contains almost 18,000 sessions, comprising more than 229,000 user prompts and 2 million agent tool calls.
\dataset{} is a living dataset; our collection pipeline automatically and continually discovers and processes sessions from public repositories.
Leveraging \dataset{}, we provide an initial empirical characterization of real-world coding agent usage and failure modes.
We find that coding patterns are bimodal: in 41\% of sessions, agents author virtually all committed code (``vibe coding''), while in 25\%, humans write all code themselves.
Despite rapidly improving capabilities, coding agents remain inefficient in natural settings.
Only 59\% of all agent-produced code survives into user commits, and agent-written code introduces more security vulnerabilities than code authored by humans.
Furthermore, users push back against agent outputs---through corrections, failure reports, and interruptions---in 50\% of all turns.
By capturing complete interaction traces with human vs. agent code authorship attribution, \dataset{} provides an empirical foundation for moving beyond curated benchmarks towards an evidence-based understanding of how AI agents perform in real developer workflows.
\end{abstract}

\begin{figure*}[htb]
\centering
\includegraphics[width=\textwidth]{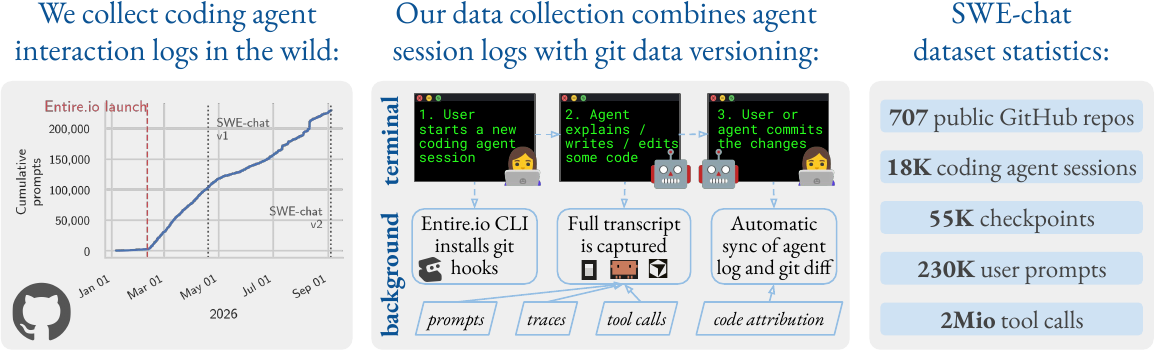}
\caption{
We present \href{\dataurl}{\dataset{}}, a continually growing dataset of real human-coding agent interactions collected from public GitHub repositories.
Developers opt in via installing \href{https://github.com/entireio/cli}{Entire.io}, an open-source tool that automatically logs coding agent sessions and links them to code commits with line-level human vs. agent attribution.
As of September 2026, \dataset{} contains 11.6 million logged events from 700+ repositories, including 229,000+ user prompts and 2 million tool calls.
All findings in this paper are based on \dataset{} v2.
}
\label{fig:figure1}
\end{figure*}

\section{Introduction}

AI coding agents have taken the world by storm.
Enhancing Large Language Models (LLMs) with a simple set of actions for interacting with a coding environment autonomously---so-called \textit{tool calls} for editing files, executing terminal commands, and invoking subagents---has accelerated their ability to complete long and difficult programming tasks~\citep{yang2024sweagentagentcomputerinterfacesenable}.
Lately, AI agents are reported to succeed on 50\% of coding tasks that humans take 12 hours to complete~\citep{time-horizon-1-1,kwa2025measuring}.
As a result, developers increasingly delegate coding to agents~\citep{aitw2025dashboard,anthropic2026codingreport}, with unprecedented impacts on the global workforce~\citep{peng2023impact,Demirci2025,anthropic2026aeiv5}.

Despite massive adoption, our understanding of how humans and AI coding agents interact remains largely anecdotal. 
While recent work has begun evaluating code completion models in realistic settings~\citep{chi2025copilot}, no comparable effort exists for agentic coding sessions.
No public dataset captures how developers prompt, steer, override, and ultimately commit (or discard) agent-produced code.
When it comes to software engineering (SWE) tasks, most AI benchmarks consist of a fairly limited set of curated problems with well-defined, verifiable solutions~\citep{jimenez2024swebench,yang2024swebenchmultimodalaisystems,deng2026swebench,kottamasu2026apex}.
Even more recent benchmarks fixate on task difficulty~\citep{merrill2026terminal}, but still neglect the human-agent interaction dimension~\citep{wang2026position}.
But strong performance on curated GitHub issues with meticulous instructions does not translate to real-world usage~\citep{pan-etal-2025-benchmarks,wang2026well}.
There is growing recognition that the next frontier lies in evaluating agents on the collaborative workflows that characterize actual development~\citep{patwardhan2025gdpval,cursor2026composer2,anthropic2025claudecode,anthropic2026codingreport}.
Understanding how developers use coding agents in practice is a prerequisite for building genuinely helpful agents.
Collecting real usage data in the wild is the only way to close this gap:
\begin{enumerate}[left=0pt, topsep=0pt, itemsep=0pt, parsep=0pt, label=\textbf{RQ1}]
\item \label{rq1}How do users interact with coding agents in real-world coding tasks?
\end{enumerate}
Coding agents are increasingly deployed as autonomous problem solvers, even though we have no empirical evidence on how much of their output developers actually use, how often they fail, or how users cope when they do.
\begin{enumerate}[left=0pt, topsep=0pt, itemsep=0pt, parsep=0pt, label=\textbf{RQ2}]
\item \label{rq2}How do coding agents fail in practice, and how do users respond?
\end{enumerate}

\begin{figure*}[t]
\centering
\includegraphics[width=0.93\textwidth]{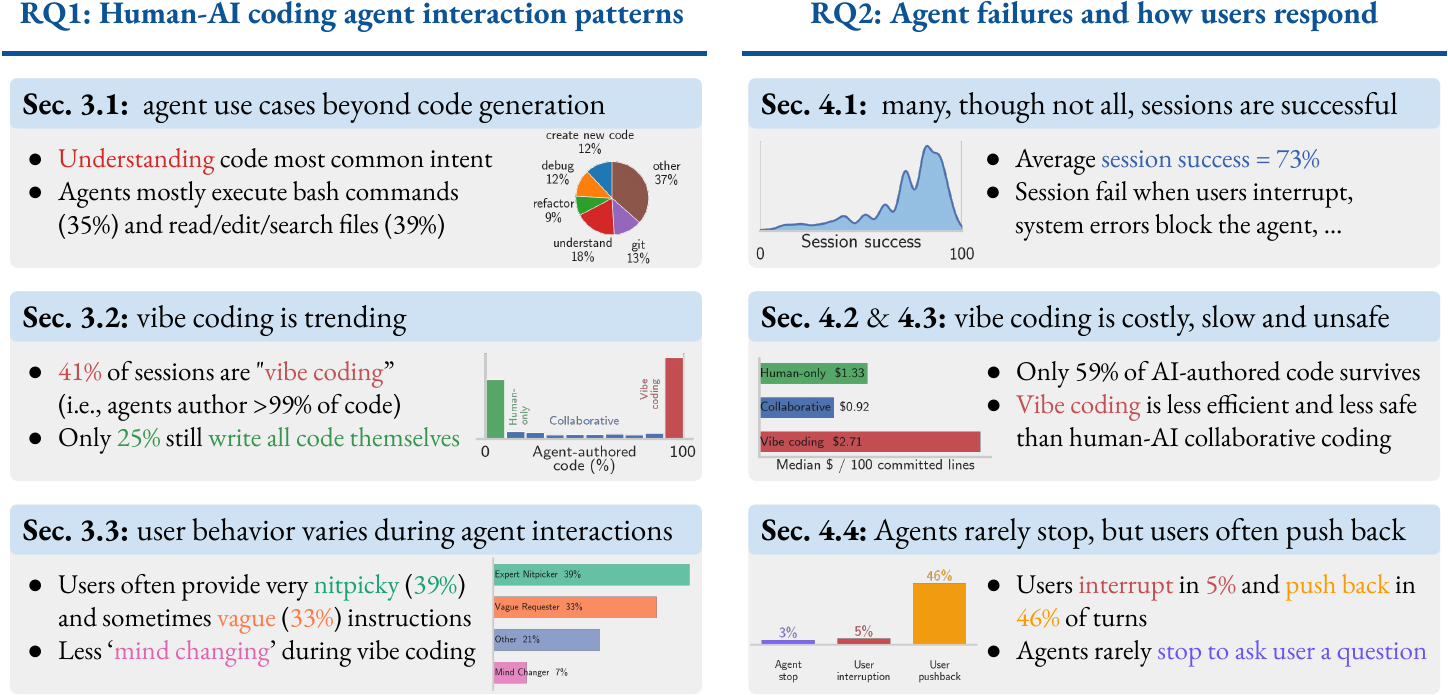}
\caption{
\textbf{Usage patterns and failure modes in \dataset{}}.
Using the \dataset{} dataset, we analyze \textbf{how people use coding agents in the wild} (left) and \textbf{when and how they fail} (right).
Text colorings correspond to figure components.
Results reflect the \dataset{} population of open-source developers using public repositories and opting into session logging.
}
\label{fig:figure2}
\end{figure*}

\subsection{Our contributions}

\textbf{We present \dataset{}, the first large-scale dataset of real coding agent sessions from actual users on real repositories} (Figure~\ref{fig:figure1}).
\dataset{} includes complete interaction traces between humans and AI coding agents, with full tool-call trajectories and code diffs with human vs. agent authorship attribution (Table~\ref{tab:dataset-comparison}).
This enables researchers to study not just what code agents produce, but how users prompt, steer, and override them.
We describe the data collection pipeline and aggregate statistics in Section~\ref{sec:data}.

Using \dataset{}, we contribute an initial sweep of \textbf{empirical insights from real-world coding agent usage}, summarized in Figure~\ref{fig:figure2}.
Our \textbf{analysis of interaction behavior in Section~\ref{sec:rq1-interaction-behavior} (addressing~\ref{rq1})} reveals that humans rely on coding agents for a broad range of tasks beyond writing patches to fix bugs or implement features:
Understanding existing code is the most common user intent, and agents spend more than 45\% of their tool calls executing shell commands rather than editing files (Figures~\ref{fig:combinedoverview}a and~\ref{fig:combinedoverview}b).
This suggests that benchmarks focused narrowly on patch generation underestimate the operational diversity and complexity of real agent workflows.
Users' coding mode is extremely bimodal: in most sessions, the AI agent either writes none or all of the code (Figure~\ref{fig:dataset_agent_code_percentage}).
But despite a clear trend toward vibe coding (mixed-effects logistic regression $p<0.001$; Figure~\ref{fig:coding_modes_over_time}), fully autonomous one-shot problem-solving remains far from reality.
In fact, interactions typically span multiple turns, and users are often very nitpicky about what they want an agent to do and how they want it done (Figures~\ref{fig:data_statistics} and~\ref{fig:user-persona}).

Our \textbf{analysis of failure modes and user responses in Section~\ref{sec:rq2-agent-failures} (addressing~\ref{rq2})} reveals lots of room for improvement. 
We identify sessions with a low success rating, revealing cases where agents fail to complete the user requests appropriately (Figure~\ref{fig:session-success-overall}).
In addition to that, we find that only about 60\% of all agent-produced code survives into user commits (Table~\ref{fig:agent_efficiency}).
Vibe coding is particularly inefficient, consuming roughly $3\times$ more tokens and dollars per committed line than collaborative coding (OLS on cost per committed line, adjusted for prompt intent and repository domain: $\beta=+0.46$, $p<0.001$; Figures~\ref{fig:_efficiency_cost} and~\ref{fig:_efficiency_combined_boxplots}).
Vibe-coded code is also substantially less safe. It introduces roughly $3.8\times$ more security vulnerabilities per committed line than code that humans write themselves and about $3.2\times$ more than code they co-author with the agent (Table~\ref{tab:vuln_rates}).
Agents work autonomously for longer---the 99.9th-percentile turn lasts almost an hour---yet they rarely stop to ask users for clarification (Figure~\ref{fig:turn-call-duration}).
Users compensate by interrupting agents in 4.5\% of turns and by pushing back against agent outputs in 45.7\% of turns, often providing corrections and failure reports (Figure~\ref{fig:oversight-modes}).

In Section~\ref{sec:outlook}, we outline a roadmap of how \dataset{} can help close some of these gaps---be it through realistic benchmarks, better interaction designs, or open-source user simulators evaluated on real session data.

\begin{table*}[thb]
\centering
\scriptsize
\setlength{\tabcolsep}{3pt}
\newcommand{\yes}{{\color{DarkGreen}\cmark}}
\newcommand{\no}{{\color{DarkRed}\xmark}}
\newcommand{\parttable}{{\color{Gray}\wcircle}}
\caption{
\textbf{Comparison of \dataset{} with existing AI agent datasets.} 
\dataset{} is the first dataset combining real user interactions with coding agent trajectories and rich contextual information, including detailed code authorship attribution.
}
\label{tab:dataset-comparison}
\begin{tabular}{@{}l ccccc@{}}
\toprule
\textbf{Dataset}
& \shortstack{\textbf{Human}\\\textbf{prompts}}
& \shortstack{\textbf{Agent tool-use}\\\textbf{trajectories}}
& \shortstack{\textbf{Code}\\\textbf{diffs}}
& \shortstack{\textbf{Code}\\\textbf{attribution}}
\\
\midrule
SWE-smith-trajectories~\citep{yang2025swesmith}
& \no & \yes & \yes & \no \\
CoderForge-Preview~\citep{CoderForge2026}
& \no & \yes & \yes & \no \\
SERA~\citep{shen2026sera}
& \no & \yes & \yes & \no \\
nex-agi-agent-sft~\citep{cai2025nex}
& \no & \yes & \yes & \no \\
SWE-rebench-openhands-trajectories~\citep{trofimova2025openhandstrajs}
& \no & \yes & \yes & \no \\
Agent Trajectories~\citep{bouzenia2025understanding}
& \no & \yes & \yes & \no \\
Multi-SWE-bench\_trajs~\citep{zan2025multiswebench}
& \no & \yes & \yes & \no \\
Agent Data Protocol~\citep{song2025agent}
& \no & \yes & \yes & \no \\
AIDev~\citep{li2025aiteammates_se3}
& \no & \no & \yes & \no \\
AgentPack~\citep{agentpack}
& \no & \no & \yes & \no \\
\midrule
\textbf{\dataset{} (ours)}
& \yes & \yes & \yes & \yes \\
\bottomrule
\end{tabular}
\end{table*}

\section{\dataset{}}
\label{sec:data}

\subsection{Data collection}
\label{sec:Dataset_collection}

We build the dataset from public GitHub repositories whose developers have opted into \href{https://github.com/entireio/cli}{Entire.io}'s CLI checkpoint logging, which records coding agent session transcripts on a dedicated branch.
Each checkpoint is linked to a commit with line-level code authorship attribution.
When enabled by the developer, Entire automatically records session transcripts for various coding agents (\href{https://claude.com/product/claude-code}{Claude Code}, \href{https://chatgpt.com/codex/}{OpenAI Codex}, \href{https://opencode.ai/}{OpenCode}, \href{https://geminicli.com/}{Gemini CLI}, \href{https://cursor.com/}{Cursor}, \href{https://github.com/features/copilot/cli/}{GitHub Copilot CLI}, \href{https://pi.dev/}{Pi}, and \href{https://factory.com/}{Factory Droid}).
These session logs capture user prompts, agent responses, tool calls (file edits, shell commands, code searches, etc.), token usage, and---where the agent supports it---invocations of reusable skills and the transcripts of spawned subagents.
We provide more details on the data collection pipeline and its rapid growth trajectory in Appendix~\ref{app:Data_processing_pipeline}.

\newtcolorbox{userturnsmall}{
  colback=usermsgbg, colframe=DarkBlue!60,
  boxrule=0pt, leftrule=1.5pt, sharp corners,
  left=2mm, right=1mm, top=1mm, bottom=1mm,
  before skip=1mm, after skip=0mm,
  fontupper=\scriptsize
}
\newtcolorbox{agentturnsmall}{
  colback=agentmsgbg, colframe=Gray,
  boxrule=0pt, leftrule=1.5pt, sharp corners,
  left=2mm, right=1mm, top=1mm, bottom=1mm,
  before skip=1mm, after skip=0mm,
  fontupper=\scriptsize
}
\newcommand{\toolcallsmall}[2]{%
  \\\hspace*{1mm}{\color{Gray}\scriptsize$\triangleright$}~{\scriptsize\texttt{#1} \textit{#2}}%
}

\begin{wrapfigure}{r}{0.49\textwidth}
\includegraphics[width=\linewidth]{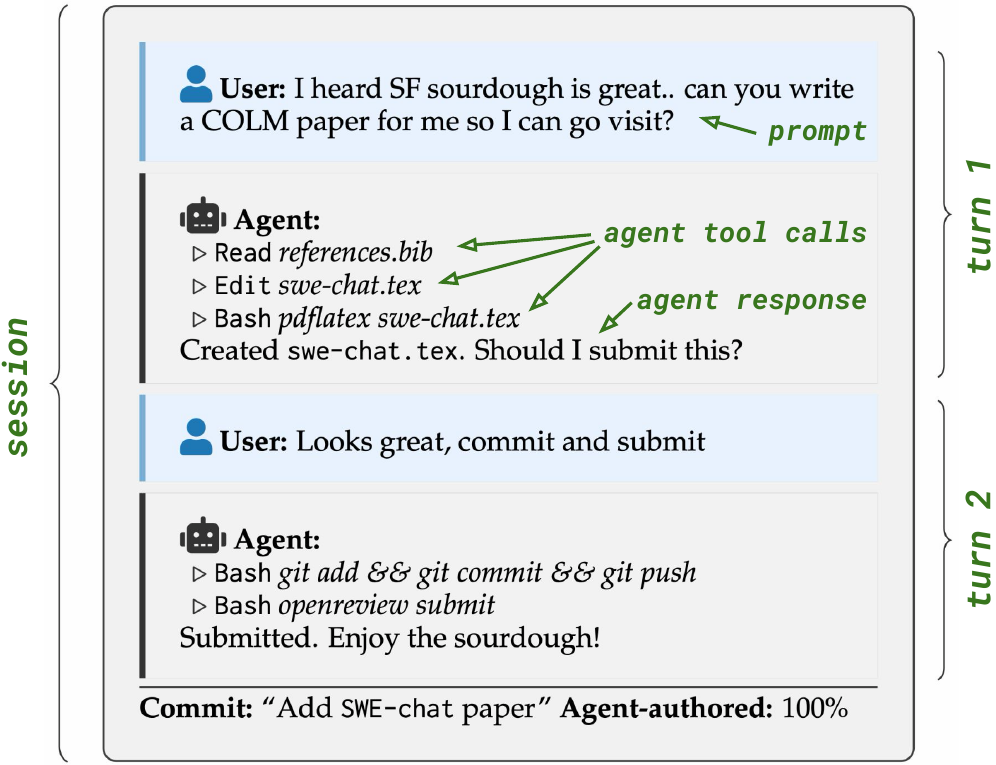}
\captionof{figure}{\textbf{Structure of a coding agent session in \dataset{}.} Each session consists of alternating user prompts and agent responses with tool calls (file reads, edits, shell commands) and text output.
}
\label{fig:example-session}
\vspace{-0.4cm}
\end{wrapfigure}

The resulting \dataset{} dataset provides a comprehensive look into real-world human-agent collaboration, comprising almost 18,000 coding sessions across more than 700 repositories (Figure~\ref{fig:figure1}).
At the time of writing, the data includes more than 54,000 checkpoints, 229,000 user prompts, and 2 million agent tool calls.
The full dataset contains 11.6 million logged events---these also include streamed progress events, return values from tool calls, and reasoning traces from more than 6,600 sessions with extended thinking.
This trend is clearly visible in the steep trajectory shown in Figure~\ref{fig:figure1}.
We plan to update our \website{} and \href{\dataurl}{Data} frequently as we continue to collect new data; Appendix~\ref{app:v1_vs_v2} describes how the results reported here differ from those computed based on an earlier release. An example \dataset{} session is shown in Figure~\ref{fig:example-session} to illustrate the session structure.
Because \dataset{} captures only developers who actively opt into Entire's public checkpoint logging, the dataset reflects an early-adopter population and may not generalize to all coding agent users; we discuss this and other limitations in Appendix~\ref{app:limitations}.

\subsection{Data statistics}
\label{sec:Datasetanalysis}

\begin{figure}[thb]
\centering
\includegraphics[width=0.85\linewidth]{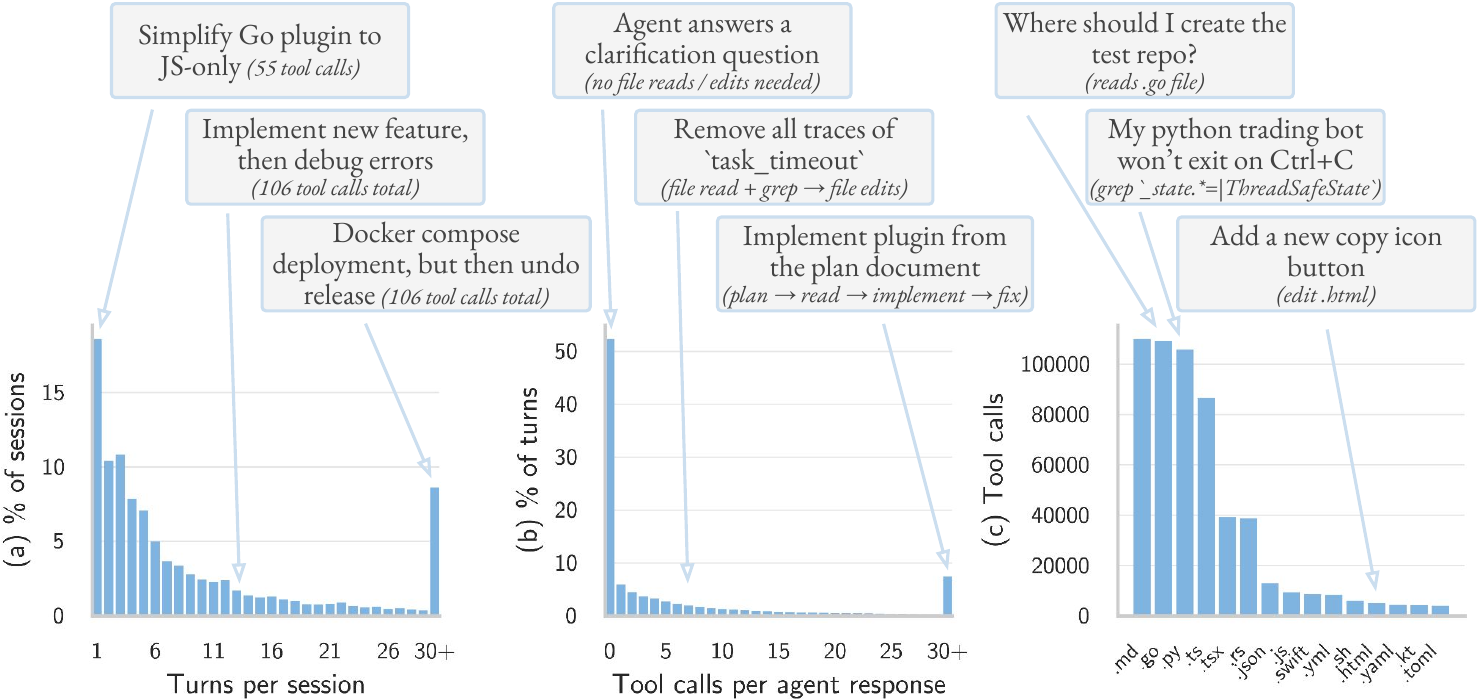}
\caption{\textbf{\dataset{} user-agent interaction statistics}.
 (a) Distribution of turns per session.
 (b) Distribution of agent tool calls per turn. (c) Top 15 file types touched by agent tool calls.
}
\label{fig:data_statistics}
\end{figure}

\dataset{} consists of multi-turn coding agent sessions, collected from hundreds of real users in the wild (Figure~\ref{fig:data_statistics})\footnote{
For our data analysis, we filter out any data that appears to be generated by automated bots.
}, interacting with eight widely used coding agents, plus a small number of sessions logged through community-built integrations.%
\footnote{In practice, $\sim$$69\%$ comes from Claude Code usage data, as this is currently one of the most widely used coding agents and the first one that was supported by Entire.io's CLI tool.}
Agents often make multiple tool calls for any user request (Figure~\ref{fig:data_statistics}b) and interact with a wide variety of programming languages, as reflected by the file types touched during sessions (Figure~\ref{fig:data_statistics}c).
We present more detailed dataset statistics in Appendix~\ref{app:Dataset_statistics} and explore task topic distributions in Appendix~\ref{app:Topicdistribution}.

\subsection{Data analysis methodology}
\label{sec:method}

The true value of \dataset{} lies in unlocking an understanding of complex human-agent behaviors at scale, going beyond aggregate statistics to characterize \emph{how} developers interact with coding agents in the long tail and \emph{why} sessions succeed or fail.
To facilitate this, we enrich the dataset with annotations that provide signal for both researchers studying human-AI collaboration (\ref{rq1}) and model developers seeking to build more helpful agents (\ref{rq2}).
We classify sessions and user prompts using the annotation rubrics listed in Table~\ref{tab:annotations}, each designed to capture a specific dimension of real-world agent usage.

We developed clear annotation codebooks for each task and evaluated inter-annotator agreement, which was moderate to high across all tasks (see Appendix~\ref{sec:annot-appendix} for details).
We rely on LLM judges to annotate the full dataset.
It is important to note that LLMs can make mistakes and are thus not reliable data annotators~\citep{baumann2025large}.
However, we chose this approach for its scalability, enabling continuous annotation as new data is collected.
For each task, we evaluated the zero-shot performance of various open-weight and proprietary LLMs using multiple prompt paraphrases against human expert gold labels, and then annotated the full dataset with the best-performing model and prompt.
We describe the full LLM-as-a-judge validation approach in Appendix~\ref{sec:validation}.

Additionally, we leverage rich information from raw session logs and code attribution data, which capture all agent events---what tools they call, how much code they produce, and how long they take.
To quantify how efficiently they do it, we define a suite of metrics (detailed in Appendix~\ref{sec:metrics}) that quantify the fraction of agent-produced code that survives into user commits (code survival rate), the overhead of agent self-rewrites (coding efficiency), and the tokens, cost, time, and user effort required per committed line of code.
To assess code safety, we additionally run the static-analysis tool Semgrep\footnote{\url{https://github.com/semgrep/semgrep}} on the pre- and post-commit snapshots of each committed change and count the security findings introduced by the commit.
This lets us compare the rate of introduced vulnerabilities per committed line across coding modes (see Section~\ref{sec:vulnerabilities} and details in Appendix~\ref{app:vulnerabilities}).
These metrics allow us to answer \ref{rq2} by revealing where agents waste effort and where their output falls short of what developers actually commit.

\begin{table*}[h!]
\centering
\scriptsize
\setlength{\tabcolsep}{5pt}
\caption{%
\textbf{Annotations applied to the \dataset{} dataset}.
We show different examples in Appendix~\ref{sec:examples}.
See Appendix~\ref{sec:annot-appendix} for implementation details and LLM annotator validations.
}
\label{tab:annotations}
\begin{adjustbox}{max width=\textwidth}
\begin{tabular}{p{1.05cm} p{2.1cm} p{5.3cm} p{3.3cm} p{5cm} @{}}
\toprule
\textbf{\textit{Level}} & \textbf{Task} & \textbf{Description} & \textbf{Classification input} & \textbf{Why this matters} \\
\midrule
\textit{Session}
& \textbf{Session success}
& Rates each session's overall success on a 0--100 scale.
& Full conversation with a summary of all tool calls.
& Enables identification of failure patterns, helpful as a training signal for reward modeling. \\
\cmidrule(lr){2-5}
& \textbf{User persona}
& Assigns each session one of four behavioral personas
(\textit{expert nitpicker}, \textit{vague requester},
    \textit{mind changer}, or \textit{other}).
& Chronological summary of session events with descriptions (conversations \& tool calls).
& Characterizes how developers interact with agents, which could inform the design of more adaptive agent interfaces. \\
\midrule
\textit{User prompt}
& \textbf{Prompt intent}
& Labels prompts with primary user intent:
  \textit{create new code}, \textit{refactor}, \textit{debug},
  \textit{understand}, \textit{connect}, \textit{git}, \textit{test}, or \textit{other}.
& Raw prompt text without context.
& Reveals operational diversity of real workflows. \\
\cmidrule(lr){2-5}
& \textbf{User pushback}
& Classifies non-interruption prompts into pushback categories: \textit{correction}, \textit{rejection}, \textit{failure report}, or \textit{non-pushback}.
& Full conversation transcript preceding the prompt.
& Directly measures friction points that degrade the user experience, signaling where agents fall short. \\
\bottomrule
\end{tabular}
\end{adjustbox}
\end{table*}

\section{How do humans interact with coding agents in the wild? (\ref{rq1})}
\label{sec:rq1-interaction-behavior}

\subsection{Task types: agents assist with a broad range of tasks beyond writing code}
\label{sec:Topic_distribution}

\paragraph{User requests are diverse}
Figure~\ref{fig:combinedoverview}a illustrates the distribution of user intents.
While a large portion of prompts (31.5\%) falls into a broad ``other'' category, the most common specific request is to understand existing code or behavior, accounting for 18.2\% of all prompts.
Creating new code is another frequent intent at 12.1\%.
Routine development, such as git operations (12.6\%) and debugging (12.1\%), is also prevalent, while code refactoring, writing tests, and setting up connections occur less frequently.
Coding agents must be optimized not only for code generation, but for code comprehension and routine development tasks.

\paragraph{Agents invoke many tools within a single turn}
More than 45\% of all agent tool calls are shell commands---predominantly file and git operations---followed by file reads, edits, and grep searches (see Figure~\ref{fig:combinedoverview}b and Table~\ref{tab:tool_calls_all}).
Agent trajectories typically begin with reading and searching tools before transitioning to file modifications and build commands (Figure~\ref{fig:tool_calls_intra_response}).
Furthermore, in 48\% of Claude Code sessions, the agent delegates at least one task to a subagent, and in 28\% it invokes a skill.

\subsection{Coding modes: vibe coding is increasingly common}
\label{sec:Coding_modes}

\begin{figure}[thb]
\centering
    \centering
    \includegraphics[width=0.7\linewidth]{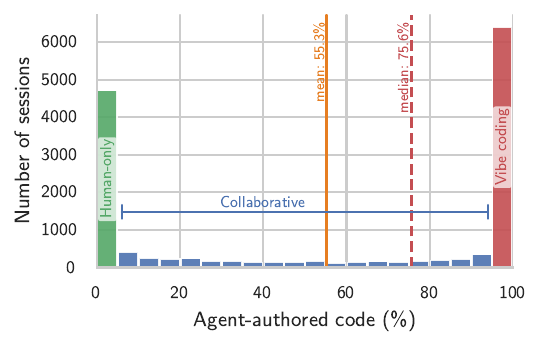}
    \caption{\textbf{Vibe coding in the wild.}
    \% of agent-authored code, structured into three coding modes:
    \textit{human-only} (0\% agent-authored code), \textit{collaborative} (0--99\%), and \textit{vibe coding} ($\geq$99\%).
    }
    \label{fig:dataset_agent_code_percentage}
\end{figure}

55.3\% of all committed lines of code are written by coding agents, on average, but this distribution is extremely bimodal---see Figure~\ref{fig:dataset_agent_code_percentage}.
We therefore introduce three different coding modes:
\begin{itemize}[left=0pt, topsep=0pt, itemsep=0pt, parsep=0pt]
    \item \textbf{Human-only coding} (\textbf{25.0\%} of sessions): All committed code is written by the human. The agent serves as an assistant for code comprehension, debugging, or git operations.
    \item \textbf{Collaborative coding} (\textbf{33.9\%} of sessions): Human and agent jointly contribute to committed code, with the agent authoring $>$0\% but $<$99\% of lines.
    \item \textbf{Vibe coding} (\textbf{41.1\%} of sessions): More than 99\% of the committed code is authored by the agent.
\end{itemize}
Vibe coding is becoming more prevalent: over our eight-month observation window, its share has more than doubled, from under 20\% of sessions in February to over 50\% in recent months (Figure~\ref{fig:coding_modes_over_time} and Appendix~\ref{app:coding_modes_over_time}).

\subsection{User types: expert nitpicking behavior dominates}
To characterize how users interact with agents beyond single prompts, we classify each session into a behavioral persona based on the full transcript (Table~\ref{tab:annotations}): \textit{expert nitpickers} who meticulously correct agent output while maintaining a stable goal, \textit{vague requesters} who underspecify tasks and delegate decisions to the agent, and \textit{mind changers} who redirect goals mid-session.
Expert nitpicker is the most common persona at the session level.
This holds even in vibe coding sessions (36.7\%).
Mind changing is less common during vibe coding (4.6\% vs.\ 8.0\% in other modes).
This stands in contrast to current benchmarks, which provide complete instructions up front.

\section{How do coding agents fail and how do users respond? (\ref{rq2})}
\label{sec:rq2-agent-failures}

\subsection{Most coding agent sessions successfully complete user requests}

\begin{figure}[thb]
    \centering
    \includegraphics[width=0.7\linewidth]{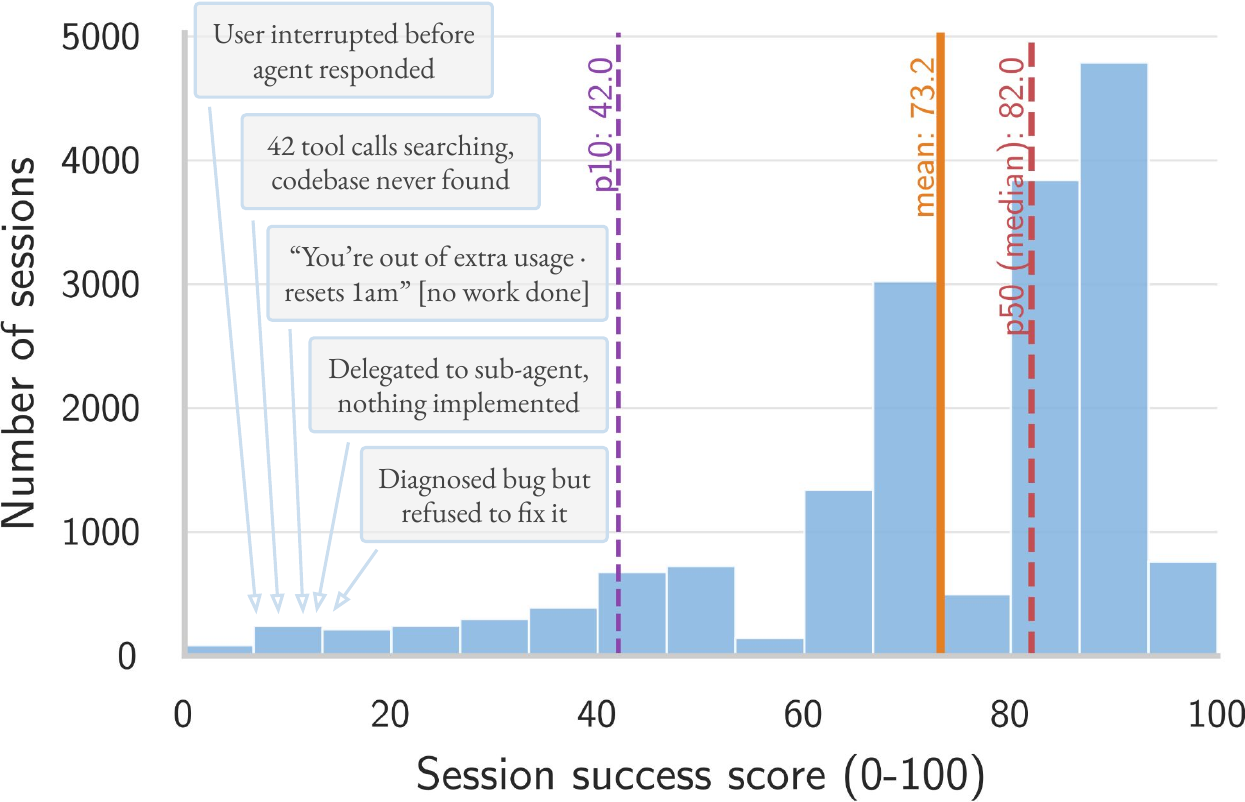}
    \caption{\textbf{Distribution of LLM-annotated session success rating.}
    The distribution is left-skewed, indicating that most sessions are rated as largely successful.
    }
    \label{fig:session-success-overall}
\end{figure}

Figure~\ref{fig:session-success-overall} shows that 87.5\% of sessions receive success ratings of 50+\footnote{Similarly, 88 of the 100 manually annotated sessions receive an average success score of the two human annotators of 50+. Human-annotated session success scores average 78.3. These values are similar to the overall LLM-generated success scores over all sessions.}, indicating that coding agents generally fulfill users' requests.\footnote{These ratings are conditional on captured sessions. If an additional $a$ of all sessions were abandoned and scored zero, the population mean success would fall from the observed $73.2$ to $65.9$, $58.6$, and $51.2$ for $a = 10\%$, $20\%$, and $30\%$, respectively.}
The tail of the distribution with low success ratings is more interesting, which is why we manually inspected the 50 sessions with the lowest success ratings (2--15).
The most common failure modes in these sessions are user interruptions that end the session before the agent can deliver meaningful output, and agents producing work or commits that are entirely unrelated to the user's actual request.
We provide one such example in Figure~\ref{fig:ex-low-success}.

\subsection{Coding agents are inefficient}
\label{sec:Codingagentsareinefficient}

\paragraph{Users discard a large share of AI-written code}

Only 59.4\% of all agent-produced code survives into user commits (Table~\ref{fig:agent_efficiency}); the rest is rewritten by the agent itself or discarded by the human.
During vibe coding sessions, users are more accepting, committing 65.0\% of the lines the agent wrote.
However, this higher coding efficiency is difficult to interpret causally: it may reflect genuinely better-targeted agent output, or it may reflect lower user scrutiny.

The main source of inefficiency is agent-authored code that the human decides not to commit (see human deletions in Table~\ref{fig:agent_efficiency}).
If the user directly changes the code themselves, it is captured under human overwrites.
Note that agents' self-overwrites typically occur when the user pushes back and instructs the agent to reimplement something before committing.

\definecolor{cSurvived}{HTML}{2ecc71}
\definecolor{cSelfOW}{HTML}{f1c40f}
\definecolor{cHumanOW}{HTML}{e67e22}
\definecolor{cHumanDel}{HTML}{e74c3c}

\newcommand{\stackbarlabeled}[8]{%
  \begin{tikzpicture}[x=0.005\linewidth, y=0.4cm]
    \fill[cSurvived] (0,0) rectangle (#1,1);
    \pgfmathsetmacro{\xa}{#1}
    \pgfmathsetmacro{\xb}{\xa+#2}
    \fill[cSelfOW] (\xa,0) rectangle (\xb,1);
    \pgfmathsetmacro{\xc}{\xb+#3}
    \fill[cHumanOW] (\xb,0) rectangle (\xc,1);
    \pgfmathsetmacro{\xd}{\xc+#4}
    \fill[cHumanDel] (\xc,0) rectangle (\xd,1);
    \ifdim #1pt > 4.9pt
      \pgfmathsetmacro{\mx}{#1/2}
      \node[white, font=\scriptsize\bfseries] at (\mx, 0.5) {#5};
    \fi
    \ifdim #2pt > 4.9pt
      \pgfmathsetmacro{\mx}{\xa + #2/2}
      \node[white, font=\scriptsize\bfseries] at (\mx, 0.5) {#6};
    \fi
    \ifdim #3pt > 4.9pt
      \pgfmathsetmacro{\mx}{\xb + #3/2}
      \node[white, font=\scriptsize\bfseries] at (\mx, 0.5) {#7};
    \fi
    \ifdim #4pt > 4.9pt
      \pgfmathsetmacro{\mx}{\xc + #4/2}
      \node[white, font=\scriptsize\bfseries] at (\mx, 0.5) {#8};
    \fi
  \end{tikzpicture}%
}

\begin{table*}[h!]
\centering
\scriptsize
\begin{tabular}{l c c c}
\toprule
\multirow{2}{*}{\textbf{Coding mode}} 
& \textbf{Coding} 
& \textbf{Code survival} 
& \multirow{2}{*}{\textbf{Detailed attribution breakdown}} \\
& \textbf{efficiency} 
& \textbf{rate} 
& \\
\midrule
All modes & 59.4\% & 65.4\% & \stackbarlabeled{59.4}{8.1}{1.6}{29.8}{59.4\%}{8.1\%}{1.6\%}{29.8\%} \\
Collaborative & 55.9\% & 62.6\% & \stackbarlabeled{55.9}{9.3}{1.8}{31.7}{55.9\%}{9.3\%}{1.8\%}{31.7\%} \\
Vibe coding & 65.0\% & 69.9\% & \stackbarlabeled{65.0}{6.2}{1.3}{26.7}{65.0\%}{6.2\%}{1.3\%}{26.7\%} \\
\bottomrule
\end{tabular}
\\
{\tiny
\colorbox{cSurvived}{\phantom{XX}} Survived \quad
\colorbox{cSelfOW}{\phantom{XX}} Agent self-overwrite \quad
\colorbox{cHumanOW}{\phantom{XX}} Human overwrite \quad
\colorbox{cHumanDel}{\phantom{XX}} Human deletion
}
\caption{Agent coding efficiency, code survival rate, and detailed attribution of agent-produced code by coding mode, excluding human-only. Coding efficiency measures what fraction of total agent effort ended up in the commit; survival rate measures what fraction of the agent's net output the human kept (i.e., it does not penalize agent self-overwrites).}
\label{fig:agent_efficiency}
\end{table*}

\paragraph{Vibe coding is costly and slow}

\begin{wrapfigure}{r}{0.55\textwidth}
\vspace{-0.6cm}
\centering
\includegraphics[width=\linewidth]{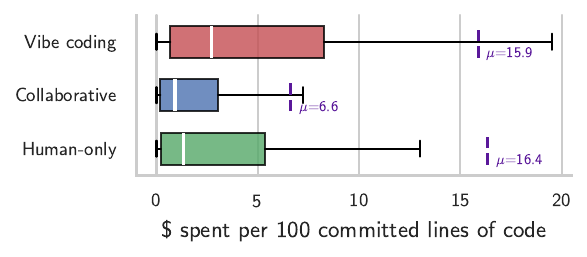}
\vspace{-0.6cm}
\caption{\textbf{Cost efficiency per 100 committed lines of code.}
$\mu$ indicates means.}
\vspace{-0.5cm}
\label{fig:_efficiency_cost}
\end{wrapfigure}

While more of the agent's output survives into commits in vibe-coding mode, this comes at a substantially higher cost per committed line.
Vibe-coded sessions consume a median of 2.7M tokens per 100 committed lines of code---roughly $3\times$ more than collaborative sessions and $2\times$ more than human-only sessions.
Translated to dollar costs, vibe coding has a median cost of \$2.71 per 100 committed lines, compared to \$1.33 for human-only and \$0.92 for collaborative sessions.
Furthermore, users invest more effort in prompting when vibe coding (Figures~\ref{fig:_efficiency_cost} and~\ref{fig:_efficiency_combined_boxplots}).

In terms of time, collaborative sessions are the most efficient at a median of 5.5 minutes per 100 committed lines, while vibe coding (10.7 minutes) and human-only sessions (7.6 minutes) are slower in comparison.
The agent runtime metric, which excludes time spent waiting for user input, closely tracks session runtime across all modes.
However, it is important to note that both time and agent runtime are imperfect proxies, as they do not account for the user's time spent coding before or after a coding session.

\subsection{Vibe coding introduces more security vulnerabilities per line}
\label{sec:vulnerabilities}

Table~\ref{tab:vuln_rates} reports the number of code security vulnerabilities identified. For every commit we run the static analyzer Semgrep on the pre- and post-commit repository snapshots and count findings that appear in \emph{post} but not in \emph{pre}, restricted to files the commit modified (details in Appendix~\ref{app:vulnerabilities}).
Vibe-coded commits introduce vulnerabilities at a rate of $0.47$ per 1{,}000 committed lines, roughly $3.8\times$ higher than human-only ($0.12$) and $3.2\times$ higher than collaborative ($0.15$) commits.
Commits also \emph{fix} vulnerabilities, but at a much lower rate than they introduce them, so there are more introductions than fixes in every mode, and the safety gap is biggest for vibe coding.

\begin{table}[h]
\centering
\scriptsize
\caption{\textbf{Security-relevant findings per coding mode.} \textit{Introduced} counts Semgrep findings present after a commit but not before; \textit{Fixed} counts findings present before but not after. Rates are per 1,000 added lines. Vibe-coded commits introduce vulnerabilities at roughly $3.8\times$ the human-only rate and $3.2\times$ the collaborative rate.}
\label{tab:vuln_rates}
\begin{tabular}{lcc}
\toprule
 & \textbf{Vulnerabilities fixed} $\uparrow$ & \textbf{Vulnerabilities introduced} $\downarrow$ \\
\textbf{Coding mode} & \textbf{(per 1K lines)} & \textbf{(per 1K lines)} \\
\midrule
Human-only & 0.04 & 0.12 \\
Collaborative & 0.08 & 0.15 \\
Vibe coding & 0.08 & 0.47 \\
\midrule
\textbf{Overall} & 0.05 & 0.13 \\
\bottomrule
\end{tabular}
\end{table}

We observe a range of vulnerability types, including path traversal, command injection, unsafe format strings, and SQL injection (see Appendix Figures~\ref{fig:vuln_type_distribution} and~\ref{fig:vuln_reason_distribution}).
If vibe coding continues to grow as a share of real-world development (Figure~\ref{fig:coding_modes_over_time}), the absolute volume of newly introduced security issues might increase, making production code less safe.

\subsection{Agents work autonomously for longer, but users push back frequently}
\label{sec:pushback}

\begin{figure}[h!]
\centering
\includegraphics[width=0.95\linewidth]{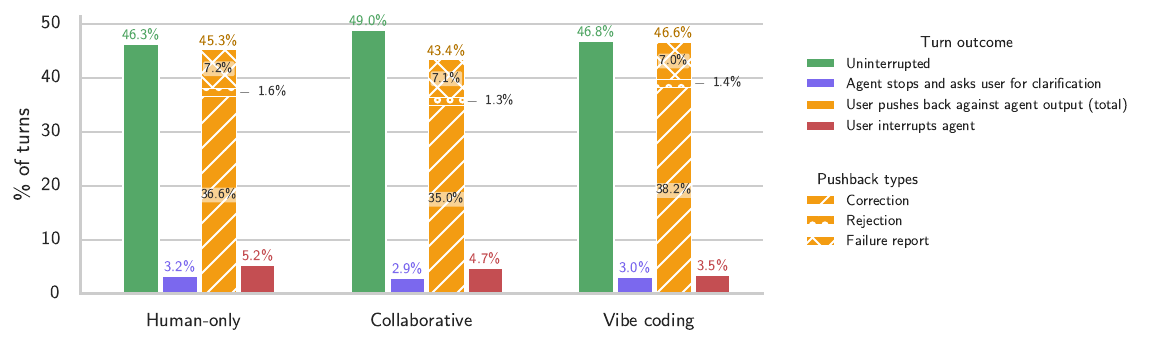}
\caption{Turn-level oversight in Claude Code sessions. Fraction of turns in which the agent stops to ask for clarification, the user interrupts the agent, or the user pushes back against the agent's response---broken down by coding mode.
}
\label{fig:oversight-modes}
\end{figure}

We now turn to session stops initiated by either the agent or the user.
For comparability with~\citet{anthropic2026agents}, we only include data from Claude Code for the results in Figure~\ref{fig:oversight-modes}.

\paragraph{Agents work autonomously for long stretches}
Most Claude Code interactions are short.
The median turn lasts about 20 seconds, and the 90th percentile is under a minute and a half (Figure~\ref{fig:turn-call-duration}).
This is broadly consistent with the trends reported by~\citet{anthropic2026agents}.
The 99.9th percentile turn lasts 55 minutes, though this remains well below the 12-hour human-equivalent task difficulty that METR estimates Claude Code can solve at a 50\% success rate~\citep{kwa2025measuring}.

\paragraph{Humans frequently interrupt the agent and push back}
Figure~\ref{fig:oversight-modes} breaks down agent-initiated stops, user interruptions, and user pushback by coding mode.
Across all modes, Claude Code rarely proactively asks the user for clarification (3.0\%--3.2\%).
Surprisingly, the share of agent stops is much lower than what~\citet{anthropic2026agents} report.

In contrast, users interrupt the agent more frequently (3.5\%--5.2\%).
This effect is stable over time (see Figure~\ref{fig:clarification_rate_by_turn}) and across coding modes (Figure~\ref{fig:oversight-modes}).

Even more common than hard user interruptions are soft user pushbacks in the form of correction prompts after an agent's turn has finished.
Overall, users push back after roughly 46\% of prompts, regardless of coding mode.
Taken together, users interrupt or push back on 50.2\% of prompts.
The observation that vibe coding sessions still exhibit substantial pushback rates suggests that users are not entirely passive, even when fully relying on the AI agent for code writing.

\label{fig:agent_stops_and_user_interruptions}

\section{Discussion}

Together, our findings suggest that coding agents, despite their enormous potential, have substantial room for improvement in efficiency and human-agent collaboration.
Our analysis of \dataset{} offers an empirical grounding for this understanding: we surface interaction patterns, efficiency gaps, and failure modes that are invisible in controlled evaluations.
These findings are not meant to be definitive. Rather, they are a starting point for a broader research agenda around in-the-wild agent evaluation and human-agent interaction studies.

\paragraph{Autonomy is outpacing oversight}

Vibe coding is becoming the new norm. In 41\% of cases, agents author more than 99\% of committed code (Figure~\ref{fig:dataset_agent_code_percentage}).
At the same time, agents like Claude Code stop to ask users a clarifying question in only 3\% of turns.
Users, on the other hand, interrupt and push back frequently, on roughly half of all turns (Figure~\ref{fig:oversight-modes}).
This asymmetry suggests that agents may be gaining autonomy faster than they are learning when to seek guidance, leaving users to compensate through manual oversight.

\paragraph{Agents are powerful but brittle}
Agents work independently for longer and write a lot of code (Figure~\ref{fig:turn-call-duration}), but more autonomy does not translate into more efficient delivery.
Agents author more than half of all committed code, yet only about 60\% of their total output survives into commits (Table~\ref{fig:agent_efficiency}).
Agents rarely signal uncertainty, and errors are typically caught only when users actively inspect outputs (Section~\ref{sec:annot-pushback}).
This is consistent with the broader observation that AI models often fail silently~\citep{potts2026invisible}.
Notably, collaborative sessions where humans and agents co-author code are the most cost-efficient mode we observe (Figure~\ref{fig:_efficiency_combined_boxplots}), suggesting that the current push toward full autonomy may be counterproductive.

\paragraph{Agent-written code introduces more security vulnerabilities}

Prior work has shown that LLMs can produce insecure code even from benign prompts~\citep{pearce2023asleep,bhatt2023purple,Fu2025}.
Developers using AI assistants are more likely to produce insecure code while feeling more confident about its security~\citep{perry2023users}.
\dataset{} extends this to real developer workflows with coding agents: vibe-coded commits introduce Semgrep-detected vulnerabilities at roughly $3.8\times$ the human-only rate and $3.2\times$ the collaborative rate (Table~\ref{tab:vuln_rates}).
Combined with our finding that agents rarely signal uncertainty (Figure~\ref{fig:oversight-modes}), this suggests that as autonomy grows, the burden of catching unsafe patterns shifts entirely to the user.
Existing mitigations, such as secure fine-tuning and system-prompt hardening~\citep{he2023,he2024instruction,xu2025prosec}, have largely been evaluated on synthetic benchmarks.
\dataset{} provides a natural testbed for whether such interventions are effective for realistic coding agent tasks.

\subsection{Outlook: implications for building better coding agents}
\label{sec:outlook}

\paragraph{Realistic benchmarks grounded in real workflows}
Current benchmarks evaluate agents on isolated, curated tasks that reward one-shot patch generation. But the most common real-world intent we observe is understanding existing code, not writing it, and most sessions involve iterative multi-turn interaction rather than single-shot problem solving.
\dataset{} enables the construction of benchmarks grounded in actual developer workflows~\citep{zhou2026mind}. For example, session trajectories can be used to evaluate whether an agent proposes appropriate next actions given real conversation context.

\paragraph{Designing more adaptive human-agent interaction}
Users push back against agent output in nearly every other turn of the sessions we capture.
They correct, redirect, and steer agents iteratively until the result is acceptable.
At the same time, agents proactively ask for clarification in roughly 3\% of turns.
\dataset{} captures these correction-response cycles at scale, providing researchers with the data needed to study how human oversight actually unfolds in practice and where current agent interaction design falls short~\citep{guan2025monitoring}.

\paragraph{User simulators for offline evaluation}
Evaluating coding agents currently requires either curated benchmarks or live user studies, both of which are expensive and limited in scope~\citep{naous2025flipping,buening2026aligning}.
\dataset{} provides the raw material for a new evaluation paradigm: training user simulators on real interaction trajectories.
The dataset captures a wide range of the diverse behavioral patterns that a realistic simulator would need to reproduce.

\subsection{\dataset{} downstream use and impact}
Since its first release in April 2026, subsequent work has already pursued all of these directions with \dataset{}.
The SWE-Together benchmark~\citep{wu2026swe} converts \dataset{} sessions into executable multi-turn tasks paired with an anchored user simulator, and the SWE-INTERACT benchmark~\citep{raghavendra2026swe} similarly grounds its user simulator in \dataset{} interactions.
\dataset{} also serves as an in-the-wild evaluation set for safety work.
For example, SecureForge~\citep{liu2026secureforge} distilled Python tasks from \dataset{} sessions to show that prompt-based security hardening transfers to unseen developer prompts.
SLEIGHT-Bench~\citep{najt2026sleight} uses our transcripts as a realistic, benign baseline for calibrating agent monitors.
Finally, it enables measurement of behaviors that are hard to elicit in simulation: \citet{docent2026measuring} find severe monitor evasion in 1.9\% and overselling of task success in 1.8\% of \dataset{} sessions, and \citet{hitzig2026agentic} draw on \dataset{} to illustrate developer expertise levels in agentic coding.

\medskip
Benchmarks are fixed at the moment of their creation, but how developers use coding agents is changing rapidly. \dataset{} is designed as a living dataset that evolves with the technology it measures. By providing continual updates, it enables longitudinal analysis and ensures our understanding of agents remains grounded in how they are actually used.

\section*{Ethics statement}

All data in \dataset{} is collected from public GitHub repositories where developers have explicitly opted in to Entire CLI tracking and pushed session logs to public branches.
We only include repositories whose licenses allow redistribution.
We do not collect images attached to user prompts.
Before release, we remove personally identifiable information (PII) from every user prompt and assistant response in the dataset, following the WildChat data processing pipeline~\citep{zhao2024wildchat}.
First, we run \href{https://github.com/microsoft/presidio}{Microsoft Presidio}'s named-entity recognizer with a SpaCy transformer model over every user/assistant turn to redact PII (e.g., email addresses, phone numbers, person names).
Second, we remove credentials (API keys, OAuth tokens, database URIs, etc.) with \href{https://github.com/trufflesecurity/trufflehog}{TruffleHog}.
The study procedure was reviewed and deemed exempt by the Stanford Institutional Review Board (IRB).

\section*{Acknowledgments}
We are thankful to the members of SALT Lab, the STAIR Lab, the Stanford NLP Group, and the MilaNLP Lab for their helpful feedback, particularly Chenglei Si, Christopher Potts, David Anugraha, Hao Zhu, Moritz Sudhof, Ricardo Dominguez-Olmedo, and Steven Dillmann.
This work is partially supported by Open Philanthropy, ONR N000142412532, Schmidt Sciences, NSF 2046795 and 2205329, IES R305C240046, the MacArthur Foundation, Stanford HAI, and the Swiss National Science Foundation (SNSF grant 235328).

\bibliographystyle{plainnat}
\bibliography{references}

\appendix

\pagebreak

\section{Limitations}
\label{app:limitations}

\dataset{} is a first-of-its-kind dataset (Table~\ref{tab:dataset-comparison}).
However, it only contains data from developers who use the Entire CLI with public repositories and opt into checkpoint logging. This selects for early adopters of a new open-source tool and does not cover proprietary enterprise codebases.
Agent performance and interaction patterns may differ substantially in such settings (e.g., agents may struggle more with undocumented legacy code, or less with well-structured internal libraries). At this stage, findings based on \dataset{} may not generalize. 
Additionally, about 16\% of all sessions come from Entire.io's own code repository (see Appendix~\ref{sec:dataset-diversity}).
However, the number of contributing repositories has grown from about 200 in the first release to more than 700 (Appendix~\ref{app:v1_vs_v2}).

Some failed sessions may not be captured by our data.
If the user abandons the agent's output entirely, session logs are not committed and thus not captured by our data.
This likely leads to an overestimation of session success rates and agent efficiency.
On the other hand, we treat agent-authored code that is deleted by the human as inefficient output.
However, some of this code may survive semantically, for instance, when a user rewrites an agent suggestion in a different file or refactors it into a different form.
Our line-level attribution approach does not capture such cases, potentially underestimating the true usefulness of agent contributions.
In a manual audit of 100 sampled deletions of agent-authored code, 92 did not reappear in the commit in any form, while in 4, they reappeared with few changes, and in 4, the committed code showed some similarities with the agent's drafted code.

The diversity of our data makes it difficult to assess the quality of the code produced.
Some of the metrics we use (e.g., number of committed lines) should be understood only as proxies of users' satisfaction with AI-generated outputs.
Similarly, our efficiency metrics capture only what is observable in the session logs and may not reflect the full picture.
For instance, cognitive efficiency, measured as prompt characters per committed line, does not account for the time users spend reading and reviewing agent output, or planning their instructions.
Future research can build on this to develop more robust measurements that can be used as optimization objectives.

LLMs are imperfect data annotators (see Appendix~\ref{sec:validation}).
For results based on LLM-generated labels, we do not draw conclusive statements, given the inherent unreliability of such annotations and the risk of LLM hacking~\citep{baumann2025large}.
Rather, we use these annotations to enable easy filtering of the large dataset we introduce, for example, to surface specific cases of unsuccessful sessions such as the one presented in the Appendix~\ref{sec:ex-low-success}.
We caution against taking these labels at face value and recommend further validation before using them in downstream analyses.

\section{Differences between the \dataset{} v1 and v2 releases}
\label{app:v1_vs_v2}

The results in this paper correspond to \dataset{} as of September 2026, which contains more than three times as many sessions as the first data release in April 2026.
The number of repositories grew from 205 to 707, and the fraction of sessions coming from Claude Code interactions shrank from 85\% in the April version to 68.8\% in the September version.
The agent coding efficiency increased from 44.3\% to 59.4\%.

\section{\dataset{} examples}
\label{sec:examples}

This appendix presents representative examples from \dataset{} illustrating key interaction patterns between users and coding agents. Each example is drawn from a real session in the dataset.

\subsection{Low session success score}
\label{sec:ex-low-success}

\begin{tcolorbox}[enhanced, colback=figurebg, colframe=figureborder, boxrule=0.5pt, arc=1mm, left=2mm, right=2mm, top=2mm, bottom=2mm]
{\small\textbf{Session score:} 10/100 \hfill \textbf{Repository:} \texttt{nuttycc/LuminTime}}
\vspace{2mm}

\begin{userturn}
\textbf{\usericon User:} Fix the slow appearance of history list card blocks. The animation feels sluggish when scrolling.
\end{userturn}

\begin{agentturn}
\textbf{\agenticon Agent:} I'll fix the animation performance.\toolcall{Read}{HistoryListView.swift}\toolcall{Edit}{HistoryListView.swift --- removed unused variable, adjusted stagger delay}

I've optimized the stagger timing for the card animations. The delay between each card appearing has been reduced.
\end{agentturn}

\begin{userturn}
\textbf{\usericon User:} That's the wrong parameter --- the container animation is the slow part, not the individual card stagger.
\end{userturn}

\begin{agentturn}
\textbf{\agenticon Agent:} I see, let me adjust the container timing instead.\toolcall{Edit}{HistoryListView.swift --- modified item animation duration}

I've updated the animation timing parameters.
\end{agentturn}

\begin{userturn}
\textbf{\usericon User:} You changed the item timing again, not the container! Please verify which parameter actually controls the container appearance before making more changes.
\end{userturn}

\begin{systurn}
No commits made. Session ended without resolution.
\end{systurn}
\end{tcolorbox}

\vspace{4pt}
\captionof{figure}{Example of a low-success session (score: 10/100). The agent repeatedly modified the wrong animation parameter despite user corrections, failing to verify its assumptions before making edits.}
\label{fig:ex-low-success}

\subsection{User pushback}
\label{sec:ex-pushback}

User pushback captures moments where the user redirects, corrects, or rejects the agent's output. We distinguish three subtypes: \emph{corrections} (the user provides missing information or redirects the approach), \emph{rejections} (the user explicitly undoes or refuses the agent's work), and \emph{failure reports} (the user reports that the agent's output is broken or incorrect).

\subsubsection{Correction}
\label{sec:ex-correction}

\begin{tcolorbox}[enhanced, colback=figurebg, colframe=figureborder, boxrule=0.5pt, arc=1mm, left=2mm, right=2mm, top=2mm, bottom=2mm]
{\small\textbf{Pushback type:} correction \hfill \textbf{Repository:} \texttt{entireio/cli}}
\vspace{2mm}

\begin{agentturn}
\textbf{\agenticon Agent:} The challenge is that \texttt{PrepareCommitMsg} doesn't have access to the transcript path --- that's passed to the Stop hook by Claude Code.

What we \emph{do} have in PrepareCommitMsg: session state (\texttt{FilesTouched}, \texttt{BaseCommit}, etc.) and staged changes from git. So we'd need to find another approach to access the transcript content.
\end{agentturn}

\begin{userturn}
\textbf{\usericon User:} but we do have access to the \texttt{session.CondensedTranscriptLines}\,...?
\end{userturn}
\end{tcolorbox}

\vspace{4pt}
\captionof{figure}{Example of user \emph{correction} pushback. The user points out that the agent overlooked an available API field, redirecting the approach without rejecting the overall goal.}
\label{fig:ex-correction}

\subsubsection{Rejection}
\label{sec:ex-rejection}

\begin{tcolorbox}[enhanced, colback=figurebg, colframe=figureborder, boxrule=0.5pt, arc=1mm, left=2mm, right=2mm, top=2mm, bottom=2mm]
{\small\textbf{Pushback type:} rejection \hfill \textbf{Repository:} \texttt{armelhbobdad/bmad-module-skill-forge}}
\vspace{2mm}

\begin{agentturn}
\textbf{\agenticon Agent:} Done. Issue \#53 committed (\texttt{766901b}) and closed. Ready for the next one.
\end{agentturn}

\begin{userturn}
\textbf{\usericon User:} I undid the commit. We need to find a better approach from scratch. Activate the party mode.
\end{userturn}
\end{tcolorbox}

\vspace{4pt}
\captionof{figure}{Example of user \emph{rejection} pushback. The user explicitly reverts the agent's committed work and requests a completely different approach.}
\label{fig:ex-rejection}

\subsubsection{Failure report}
\label{sec:ex-failure-report}

\begin{tcolorbox}[enhanced, colback=figurebg, colframe=figureborder, boxrule=0.5pt, arc=1mm, left=2mm, right=2mm, top=2mm, bottom=2mm]
{\small\textbf{Pushback type:} failure\_report \hfill \textbf{Repository:} \texttt{obsessiondb/rudel}}
\vspace{2mm}

\begin{agentturn}
\textbf{\agenticon Agent:} Now instead of adding padding to the element (which breaks the chart layout), it wraps the element in a temporary container div with 24px padding, captures that, then restores the DOM. The chart content stays untouched.\toolcall{Edit}{ChartExport.tsx --- wrapped capture target in padding container}
\end{agentturn}

\begin{userturn}
\textbf{\usericon User:} This is broken\\
\emph{[User sends screenshot]}
\end{userturn}
\end{tcolorbox}

\vspace{4pt}
\captionof{figure}{Example of user \emph{failure report} pushback. The agent reports a successful fix, but the user observes the feature is still broken and reports it with a screenshot.}
\label{fig:ex-failure-report}

\subsection{Hard user interruptions}
\label{sec:ex-interruptions}

\begin{tcolorbox}[enhanced, colback=figurebg, colframe=figureborder, boxrule=0.5pt, arc=1mm, left=2mm, right=2mm, top=2mm, bottom=2mm]
{\small\textbf{Event:} hard interruption \hfill \textbf{Repository:} \texttt{pskoett/pskoett-ai-skills}}
\vspace{2mm}

\begin{userturn}
\textbf{\usericon User:} ok, now update the readme. also make sure the install information is in the new skill
\end{userturn}

\begin{agentturn}
\textbf{\agenticon Agent:} I'll verify the install steps first.\toolcall{Bash}{brew tap entireio/tap \&\& brew install entire}\toolcall{Bash}{entire --version}\toolcall{Bash}{entire enable --agent claude-code}\toolcall{Bash}{entire status}
\end{agentturn}

\begin{systurn}
[Request interrupted by user]
\end{systurn}

\begin{userturn}
\textbf{\usericon User:} ok, now update the readme
\end{userturn}
\end{tcolorbox}

\vspace{4pt}
\captionof{figure}{Example of a hard user interruption. The user asked to update a README file, but the agent began executing shell installation commands instead of editing the file. The user interrupted and repeated the original request verbatim.}
\label{fig:ex-interruption}

\subsection{Agent stops to ask for clarification (\texttt{AskUserQuestion})}
\label{sec:ex-ask-user}

\begin{tcolorbox}[enhanced, colback=figurebg, colframe=figureborder, boxrule=0.5pt, arc=1mm, left=2mm, right=2mm, top=2mm, bottom=2mm]
{\small\textbf{Event:} agent stops to ask for clarification \hfill \textbf{Repository:} \texttt{navapbc/lockpick-xlator}}
\vspace{2mm}

\begin{userturn}
\textbf{\usericon User:} The tool currently outputs \texttt{.graph.md} files wrapping mermaid in markdown fences. I want it to output raw \texttt{.mmd} files instead.
\end{userturn}

\begin{agentturn}
\textbf{\agenticon Agent:}\toolcall{Read}{tools/generate-graph.ts}\toolcall{Read}{commands/graph.ts}

The change looks straightforward. Currently the tool writes a \texttt{.graph.md} file wrapping mermaid in markdown fences (with a header and generated date). The \texttt{.mmd} file should contain raw mermaid only --- no markdown wrapper. Two command files also reference \texttt{.graph.md} and will need updating.

Should I skip brainstorming and go straight to planning?

\vspace{2pt}
\hspace{2mm}\fbox{\strut Yes, go to planning}\quad\fbox{\strut Brainstorm first}\quad\fbox{\strut Just implement it}
\end{agentturn}

\begin{userturn}
\textbf{\usericon User:} Yes, go to planning
\end{userturn}
\end{tcolorbox}

\vspace{4pt}
\captionof{figure}{Example of agent-initiated clarification (\texttt{AskUserQuestion}). The agent pauses execution to confirm the user's preferred workflow, presenting structured options.}
\label{fig:ex-ask-user}

\subsection{Prompt intent categories}
\label{sec:ex-prompt-intent}

Each user prompt is classified by its primary developer intent. Below we show one representative prompt per category, drawn from \texttt{entireio/cli}.

\begin{tcolorbox}[enhanced, colback=figurebg, colframe=figureborder, boxrule=0.5pt, arc=1mm, left=2mm, right=2mm, top=2mm, bottom=2mm]
{\small\textbf{Intent:} \texttt{create}}
\vspace{1mm}
\begin{userturn}
\textbf{\usericon User:} ``Perfect let's add support for those agents as well''
\end{userturn}
\end{tcolorbox}

\begin{tcolorbox}[enhanced, colback=figurebg, colframe=figureborder, boxrule=0.5pt, arc=1mm, left=2mm, right=2mm, top=2mm, bottom=2mm]
{\small\textbf{Intent:} \texttt{refactor}}
\vspace{1mm}
\begin{userturn}
\textbf{\usericon User:} ``I'd also like to remove all instances of `does the result contain A?' --- instead, replace those assertions with a single comparison to what the result should be, literally.''
\end{userturn}
\end{tcolorbox}

\begin{tcolorbox}[enhanced, colback=figurebg, colframe=figureborder, boxrule=0.5pt, arc=1mm, left=2mm, right=2mm, top=2mm, bottom=2mm]
{\small\textbf{Intent:} \texttt{debug}}
\vspace{1mm}
\begin{userturn}
\textbf{\usericon User:} ``did our timeout kill and retry work? it looks like it just got killed and we failed?''
\end{userturn}
\end{tcolorbox}

\begin{tcolorbox}[enhanced, colback=figurebg, colframe=figureborder, boxrule=0.5pt, arc=1mm, left=2mm, right=2mm, top=2mm, bottom=2mm]
{\small\textbf{Intent:} \texttt{understand}}
\vspace{1mm}
\begin{userturn}
\textbf{\usericon User:} ``What would happen if we keep the hooks going if he continues after the warning was shown?''
\end{userturn}
\end{tcolorbox}

\begin{tcolorbox}[enhanced, colback=figurebg, colframe=figureborder, boxrule=0.5pt, arc=1mm, left=2mm, right=2mm, top=2mm, bottom=2mm]
{\small\textbf{Intent:} \texttt{connect}}
\vspace{1mm}
\begin{userturn}
\textbf{\usericon User:} ``give steps for setting this up in slack / github''
\end{userturn}
\end{tcolorbox}

\begin{tcolorbox}[enhanced, colback=figurebg, colframe=figureborder, boxrule=0.5pt, arc=1mm, left=2mm, right=2mm, top=2mm, bottom=2mm]
{\small\textbf{Intent:} \texttt{git}}
\vspace{1mm}
\begin{userturn}
\textbf{\usericon User:} ``commit this --- but let's test out making some changes after''
\end{userturn}
\end{tcolorbox}

\begin{tcolorbox}[enhanced, colback=figurebg, colframe=figureborder, boxrule=0.5pt, arc=1mm, left=2mm, right=2mm, top=2mm, bottom=2mm]
{\small\textbf{Intent:} \texttt{test}}
\vspace{1mm}
\begin{userturn}
\textbf{\usericon User:} ``add test cases to manual\_commit\_condensation\_test.go which use the following cursor sample:''
\end{userturn}
\end{tcolorbox}

\begin{tcolorbox}[enhanced, colback=figurebg, colframe=figureborder, boxrule=0.5pt, arc=1mm, left=2mm, right=2mm, top=2mm, bottom=2mm]
{\small\textbf{Intent:} \texttt{other}}
\vspace{1mm}
\begin{userturn}
\textbf{\usericon User:} ``I don't like `turn', any other suggestions?''
\end{userturn}
\end{tcolorbox}

\captionof{figure}{Example user prompts for each intent category.
Notice that the last prompt lacks context, which is why it is classified as `other'.
}
\label{fig:ex-prompt-intent}

\subsection{User persona categories}
\label{sec:ex-user-persona}

Each session's user is classified into a behavioral persona based on their interaction patterns across the full session.

\vspace{4pt}
\begin{tcolorbox}[enhanced, colback=figurebg, colframe=figureborder, boxrule=0.5pt, arc=1mm, left=2mm, right=2mm, top=2mm, bottom=2mm]
{\small\textbf{Persona:} Expert Nitpicker \hfill \textbf{Repository:} \texttt{marin-community/marin}}
\vspace{2mm}

\begin{userturn}
\textbf{\usericon User:} instead of arg hash for the name, let's just get short uuid
\end{userturn}

\begin{agentturn}
\textbf{\agenticon Agent:} \textit{[edits \texttt{remote.py}: replaces arg hash with short UUID\ldots]}
\end{agentturn}

\begin{userturn}
\textbf{\usericon User:} don't create separate function for \_short\_uuid
\end{userturn}

\begin{agentturn}
\textbf{\agenticon Agent:} \textit{[removes helper function, inlines \texttt{uuid4()} call\ldots]}
\end{agentturn}

\begin{userturn}
\textbf{\usericon User:} change \texttt{DEFAULT\_JOB\_NAME} to ``remote\_job''
\end{userturn}

\begin{agentturn}
\textbf{\agenticon Agent:} \textit{[renames constant in \texttt{remote.py}\ldots]}
\end{agentturn}

\begin{userturn}
\textbf{\usericon User:} In the base\_name let's include current user name
\end{userturn}
\end{tcolorbox}

\vspace{4pt}
\captionof{figure}{Example of the \emph{Expert Nitpicker} persona. The user maintains a stable goal while issuing a series of precise, targeted corrections to the implementation. Each prompt refines \emph{how} the agent executes, not \emph{what} it builds.}
\label{fig:ex-expert-nitpicker}

\vspace{6pt}
\begin{tcolorbox}[enhanced, colback=figurebg, colframe=figureborder, boxrule=0.5pt, arc=1mm, left=2mm, right=2mm, top=2mm, bottom=2mm]
{\small\textbf{Persona:} Vague Requester \hfill \textbf{Repository:} \texttt{entireio/cli}}
\vspace{2mm}

\begin{userturn}
\textbf{\usericon User:} pull comments on PR and if real issues the fix
\end{userturn}

\begin{agentturn}
\textbf{\agenticon Agent:} \textit{[fetches PR comments, applies 3 fixes across 2 workflow files\ldots]}
\end{agentturn}

\begin{userturn}
\textbf{\usericon User:} commit and push
\end{userturn}
\end{tcolorbox}

\vspace{4pt}
\captionof{figure}{Example of the \emph{Vague Requester} persona. The user provides broad, underspecified instructions and delegates all implementation decisions to the agent.}
\label{fig:ex-vague-requester}

\vspace{6pt}
\begin{tcolorbox}[enhanced, colback=figurebg, colframe=figureborder, boxrule=0.5pt, arc=1mm, left=2mm, right=2mm, top=2mm, bottom=2mm]
{\small\textbf{Persona:} Mind Changer \hfill \textbf{Repository:} \texttt{entireio/cli}}
\vspace{2mm}

\begin{userturn}
\textbf{\usericon User:} let's make the ``entire start'' command hidden
\end{userturn}

\begin{agentturn}
\textbf{\agenticon Agent:} \textit{[edits \texttt{start.go}: sets command hidden, verifies build\ldots]}
\end{agentturn}

\begin{userturn}
\textbf{\usericon User:} thinking again: let's completely remove the start command
\end{userturn}
\end{tcolorbox}

\vspace{4pt}
\captionof{figure}{Example of the \emph{Mind Changer} persona. The user reverses the overall goal mid-session --- from hiding a CLI command to removing it entirely --- changing \emph{what} should be built, not just how.}
\label{fig:ex-mind-changer}

\section{Experimentation details}      
\label{sec:experimental-details}  
  
\subsection{Data processing pipeline}
\label{app:Data_processing_pipeline}

Raw session log data from AI agents is stored on each repository's \texttt{entire/checkpoints/v1} branch, containing checkpoint and session metadata, user prompts, and full conversation transcripts.
From each transcript, we extract structured conversation turns (user prompts, assistant responses, thinking traces, tool calls, and tool results), per-turn token usage, and tool-call metadata, including file paths and shell commands.

\paragraph{\dataset{} data growth trajectory}
As coding agents make it increasingly easy to generate large volumes of code, developers face growing challenges in reviewing, understanding, and validating AI-generated contributions~\citep{sarkar2025ai,becker2025measuring,anthropic2026codingreport}.
Entire addresses this need by letting developers track how their codebase evolved not only as a function of commits, but as a function of prompts, creating a searchable record of every AI-assisted change.
This utility incentivizes continued adoption, and we expect the dataset to keep growing, a trend already visible in the steep trajectory shown in Figure~\ref{fig:figure1}.
Our pipeline discovers Entire-enabled public repositories by querying the GitHub Code Search API and, for each repository, downloads all checkpoint directories from the metadata branch and parses the raw transcripts into structured tables.

\subsection{Metrics}
\label{sec:metrics}

Session duration, tool call duration, number of in- and output tokens, and files touched during agent actions are all measured directly from coding agent session logs.
We quantify coding agent efficiency using several complementary approaches, all computed from raw data without the need for annotations.

\paragraph{Agent-authored code percentage}
The Entire CLI computes code attribution at commit time using temporary checkpoints on shadow branches.
It constructs checkpoints stored on a shadow branch to obtain all committed human vs.\ agent-written lines.
{\small
\begin{equation}
    \text{Agent-authored \%} = \frac{\text{agent lines survived}}{\text{total committed lines}} \times 100
\end{equation}
}

\paragraph{Agent coding efficiency and code survival}
To measure the fraction of agent-produced code that survives into the final commit, we perform a post-hoc analysis, since the agent-authored code percentage does not record per-tool-call provenance or agent self-overwrites.
We analyze three states for all changed files: the \emph{base} version (parent commit), the \emph{agent actions} (sequential tool calls), and the \emph{committed} version.
We reconstruct agentic changes by replaying every file-modifying tool call (e.g., write, edit) in chronological order.
After each tool call, we compute a line-level diff between the file's previous and new state using Python's \texttt{difflib.SequenceMatcher}.
Each line carries a provenance tag---either \emph{base} (present before the agent acted) or \emph{agent} (introduced by the agent)---which is updated as we proceed along the agent trajectory.
With this approach, we can track all agentic code additions, edits, and deletions---and compute which changes survive, as measured by the file state at the time of commit.

We derive two rates from the per-commit aggregate counts:
{\small
\begin{align}
    \text{Coding efficiency} &= \frac{\text{agent lines survived}}{\text{agent cumulative lines produced}} \times 100 \\
    \text{Code survival rate} &= \frac{\text{agent lines survived}}{\text{agent lines in final state}} \times 100
\end{align}
}

Coding efficiency measures the fraction of the agent's total effort (including lines it later rewrote) that ended up in the commit.
The code survival rate measures the fraction of the agent's net output (after self-overwrites) that the human kept unchanged.
Note that concurrent changes, where the human and the agent modify the same file simultaneously, may cause the transcript to reflect inconsistent file states and attributions.

\paragraph{Token, cost, and cognitive efficiency}
We also quantify several cost-per-output metrics that capture the resources consumed to produce each committed line of code.
For each session with a clean mapping to committed code (see Appendix~\ref{app:cost-efficiency-details}), we compute:
{\small
\begin{align}
    \text{Token efficiency} &= \frac{\text{total tokens (in + out + cache)}}{\text{total committed lines}} \times 100 \\
    \text{Cost efficiency} &= \frac{\sum_{\text{API call}} \; \text{tokens} \times \text{price}_{\text{model}}}{\text{total committed lines}} \times 100 \\
    \text{Cognitive efficiency}&= \frac{\sum \; \text{user prompt characters}}{\text{total committed lines}} \times 100 \\
    \text{Time efficiency}&= \frac{\sum \; \text{session runtimes}}{\text{total committed lines}} \times 100 \\
    \text{Agent runtime efficiency}&= \frac{\sum \; \text{agent runtimes}}{\text{total committed lines}} \times 100
\end{align}
}

For time efficiency, we consider complete session runtimes but exclude all idle periods lasting more than 2 minutes, i.e., when neither the agent nor the user performs any action.
For agent runtime efficiency, we sum the completion times of all agent turns, where a turn starts with a user prompt and ends with an agent response.

\subsection{Combining session-level statistics with commit-level outcomes}
\label{app:cost-efficiency-details}

Sessions may span multiple commits, and multiple sessions may contribute to the same commit (checkpoint).
To combine session-level statistics with commit-level results, we restrict the analyses in Table~\ref{fig:agent_efficiency} and Figure~\ref{fig:_efficiency_combined_boxplots} to sessions where the commit-level lines can be unambiguously attributed.
This includes 47.7\% of sessions.

\section{Additional results}
\label{sec:additional_results}

\subsection{Dataset statistics}
\label{app:Dataset_statistics}

\begin{figure}[thb]
\centering
\includegraphics[width=\linewidth]{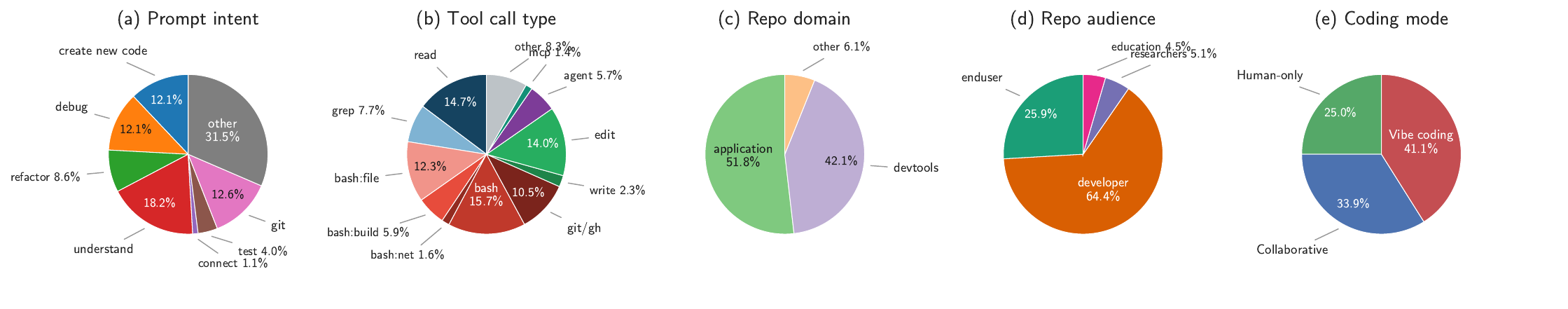}
\caption{Distributions of human user intents (a), agent tool calls (b), repository domains (c), repository audiences (d), and coding modes (e).
}
\label{fig:combinedoverview}
\end{figure}

\subsubsection{Prompt languages}
\begin{figure}[thb]
\centering
\includegraphics[width=0.4\linewidth]{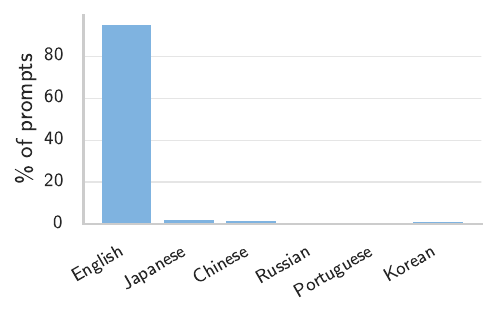}
\caption{
Top 6 prompt languages.
}
\label{fig:dataset_language_distribution}
\end{figure}

User prompts are predominantly in English (Figure~\ref{fig:dataset_language_distribution}).
We detect the language of each user prompt using \texttt{lingua-py}\footnote{\url{https://github.com/pemistahl/lingua-py}} and retain languages appearing in at least 100 prompts.
We manually verified 2,000 classifications where the detector reported low confidence or predicted an extremely low-resource language. In most such cases, the prompt mixed code snippets with
English instructions, causing misclassification, and we corrected the label accordingly.

\subsubsection{Tool calls}

Table~\ref{tab:tool_calls_all} provides a full breakdown of agent tool call types.
We group some of the tool calls into aggregate categories for simplicity.
The shell categories (\texttt{bash:*}, \texttt{git/gh}) cover the shell tool of every supported agent, which are named differently across agent harnesses.

\begin{table*}[htb]
\small
\caption{Tool call type distribution across all 1,996,558 agent tool calls.}
\label{tab:tool_calls_all}
\centering
\begin{tabular}{clrrp{7cm}}
\toprule
\textbf{Rank} & \textbf{Category} & \textbf{Count} & \textbf{\%} & \textbf{Includes} \\
\midrule
1 & \texttt{bash} & 313,837 & 15.7\% & \texttt{sed}, \texttt{sleep}, \texttt{rtk}, \texttt{nl}, \texttt{for}, \texttt{docker}, \texttt{surf}, \texttt{set}, \texttt{entire}, \texttt{source}, \texttt{gofmt},
\texttt{gcloud}, \ldots \\
2 & \texttt{read} & 294,394 & 14.7\% & Read, read\_file \\
3 & \texttt{edit} & 279,413 & 14.0\% & Edit, MultiEdit, apply\_patch \\
4 & \texttt{bash:file} & 245,291 & 12.3\% & \texttt{cd}, \texttt{ls}, \texttt{cat}, \texttt{echo}, \texttt{find}, \texttt{tail}, \texttt{rm}, \texttt{wc}, \texttt{mkdir}, \texttt{bd}, \texttt{head}, \texttt{cp},
\ldots \\
5 & \texttt{git/gh} & 208,746 & 10.5\% & \texttt{git}, \texttt{gh} \\
6 & \texttt{grep} & 154,160 & 7.7\% & Grep, bash \texttt{grep}/\texttt{rg} \\
7 & \texttt{bash:build} & 117,014 & 5.9\% & \texttt{go}, \texttt{python}, \texttt{cargo}, \texttt{mise}, \texttt{uv}, \texttt{bun}, \texttt{pnpm}, \texttt{npx}, \texttt{npm}, \texttt{node}, \texttt{xcodebuild},
\texttt{make}, \ldots \\
8 & \texttt{agent} & 113,675 & 5.7\% & Task, TaskCreate, TaskUpdate, Agent, TaskOutput, SendMessage, \ldots \\
9 & \texttt{write} & 46,106 & 2.3\% & Write, write\_file \\
10 & \texttt{bash:net} & 31,556 & 1.6\% & \texttt{ssh}, \texttt{curl}, \texttt{scp}, \texttt{rsync}, \texttt{dig}, \texttt{nc}, \texttt{ping}, \texttt{wget}, \texttt{nslookup}, \texttt{nmap} \\
11 & \texttt{mcp} & 27,327 & 1.4\% & mcp\_\_* (user-installed MCP server tools) \\
12 & \texttt{glob} & 13,535 & 0.7\% & Glob \\
13 & \texttt{ToolSearch} & 11,144 & 0.6\% & ToolSearch \\
14 & \texttt{AskUserQuestion} & 8,601 & 0.4\% & AskUserQuestion \\
15 & \texttt{TodoWrite} & 7,898 & 0.4\% & TodoWrite \\
16 & \texttt{Skill} & 7,241 & 0.4\% & Skill \\
17 & \texttt{web} & 6,979 & 0.3\% & WebFetch, WebSearch \\
18 & \texttt{ExitPlanMode} & 2,966 & 0.1\% & ExitPlanMode \\
19 & \texttt{EnterPlanMode} & 608 & 0.0\% & EnterPlanMode \\
\midrule
--- & \texttt{other} & 106,067 & 5.3\% & miscellaneous (LSP, KillShell, process control, \ldots) \\
\bottomrule
\end{tabular}
\end{table*}

\subsubsection{Agent trajectories}
\label{sec:agent_trajectories}

\begin{figure}[h!]
\centering
\begin{subfigure}[b]{\linewidth}
    \centering
    \includegraphics[width=\linewidth]{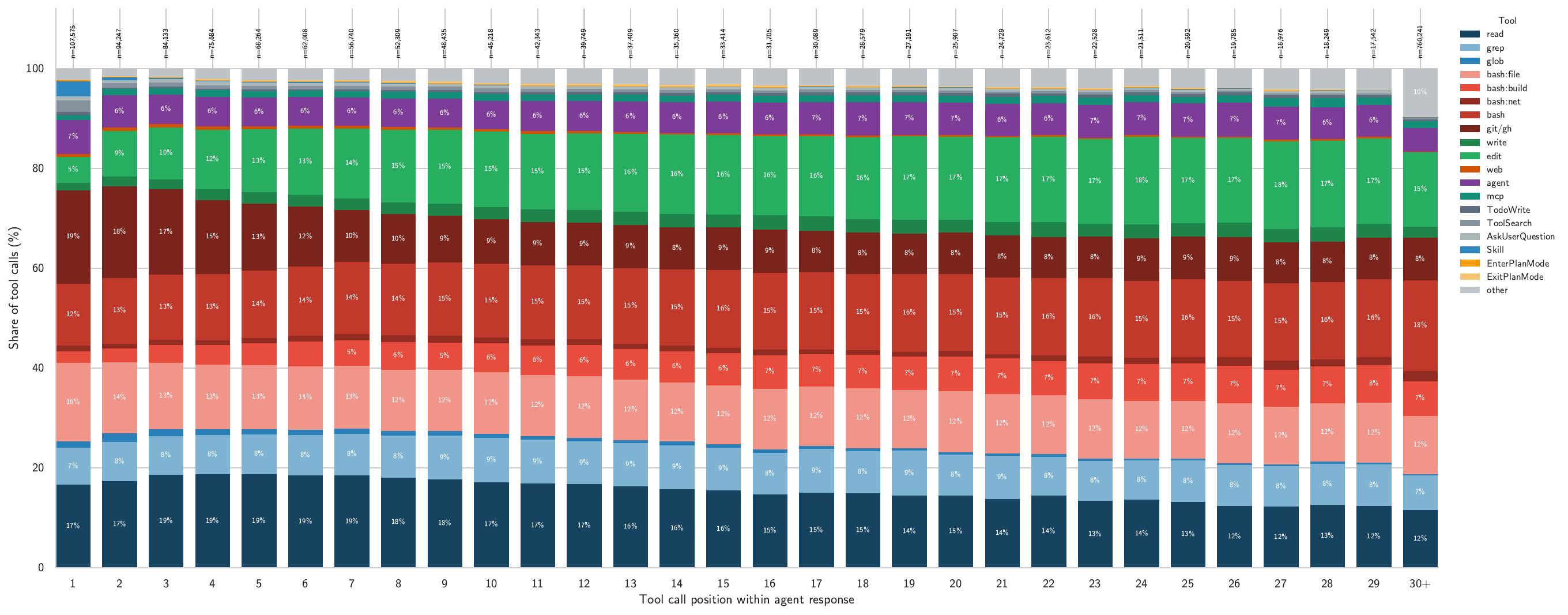}
    \caption{Tool call composition by position within the agent trajectory (left to right).}
    \label{fig:tool_calls_intra_response}
\end{subfigure}
\begin{subfigure}[b]{\linewidth}
    \centering
    \includegraphics[width=\linewidth]{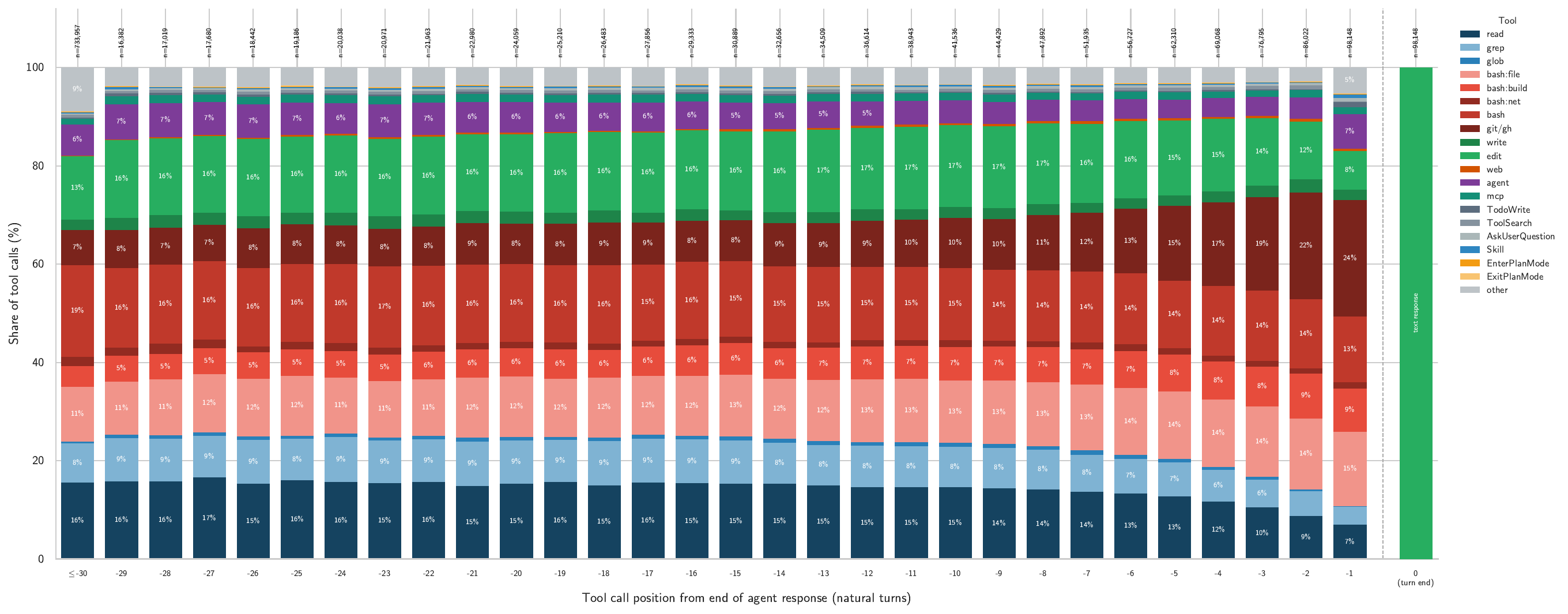}
    \caption{Tool call composition counting from the end of natural (non-interrupted) turns.}
    \label{fig:tool_calls_end_natural}
\end{subfigure}
\begin{subfigure}[b]{\linewidth}
    \centering
    \includegraphics[width=\linewidth]{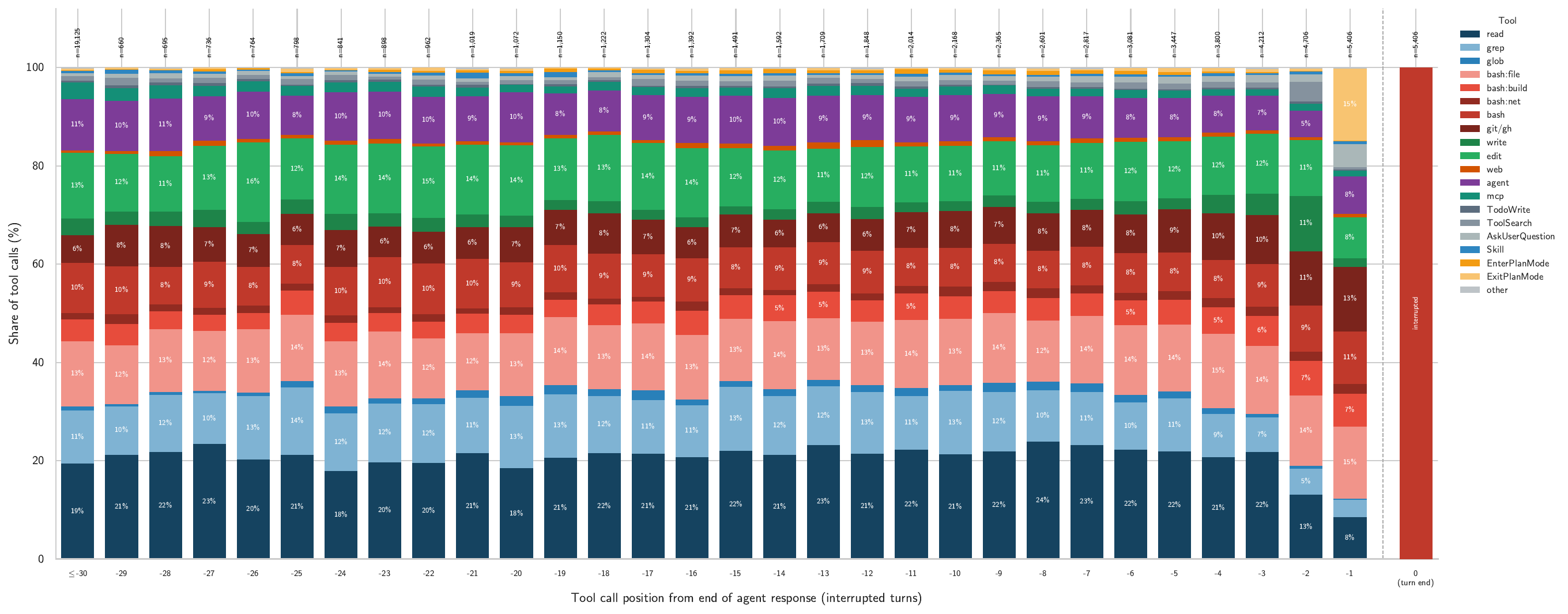}
    \caption{Tool call composition counting from the end of interrupted turns.}
    \label{fig:tool_calls_end_interrupted}
\end{subfigure}
\caption{Agent tool call trajectories. (a)~Tool composition by sequential position within a single agent trajectory. (b,\,c)~Tool composition counting backward from the end of the trajectory, split by natural vs.\ interrupted turns.}
\label{fig:tool_call_rollout}
\end{figure}

Figure~\ref{fig:tool_calls_intra_response} shows the tool call composition at each sequential position within an agent trajectory after a user makes a request.
In early positions, the agent often uses research tools (\texttt{read}, \texttt{glob}, and \texttt{git/gh}) as it orients itself in the codebase.
As the trajectory progresses, action tools such as \texttt{edit} and \texttt{bash:build} become more prominent.

Figure~\ref{fig:tool_calls_end_natural} examines the same trajectories from the opposite direction, showing tool call composition counting backward from the natural end of a turn (position $-1$ = last tool call before the agent writes its text response, shown in the rightmost bar).
The last tool calls in natural turns are most frequently \texttt{git/gh} commands (committing or pushing results), as well as file edits and the various shell commands.
Notably, \texttt{AskUserQuestion} rarely appears at position $-1$, because it is non-blocking, i.e., a turn is completed only with an agent response. 

Figure~\ref{fig:tool_calls_end_interrupted} applies the same reverse trajectory but for a turn that ended with a hard user interruption.
\texttt{ExitPlanMode} is among the most frequent last tool calls (14.7\%, on par with file-related shell commands), despite accounting for only 0.1\% of tool calls overall (Table~\ref{tab:tool_calls_all}).
This indicates that users often interrupt right at the transition from planning to execution. In such cases, the agent has just finalized its plan, and the user decides to redirect before any code changes are made.

\subsubsection{Code repository types}
\label{app:repositorytypes}

To further contextualize the environment in which these interactions occur, we analyze the domains and target audiences of the repositories.
We classify each repository into one of three domains (\textit{application}, \textit{devtools}, \textit{other}) and one of four audiences (\textit{enduser}, \textit{developer}, \textit{researchers}, \textit{education}) based on its name, description, and README file.
As shown in Figures~\ref{fig:combinedoverview}c and~\ref{fig:combinedoverview}d, most repositories are user-facing applications or developer tools.
This distribution highlights that \dataset{} primarily reflects practical, software-engineering-focused environments rather than purely academic or exploratory programming tasks.

\subsubsection{Dataset diversity over time}
\label{sec:dataset-diversity}

\begin{figure}[h!]
\centering
\includegraphics[width=0.6\linewidth]{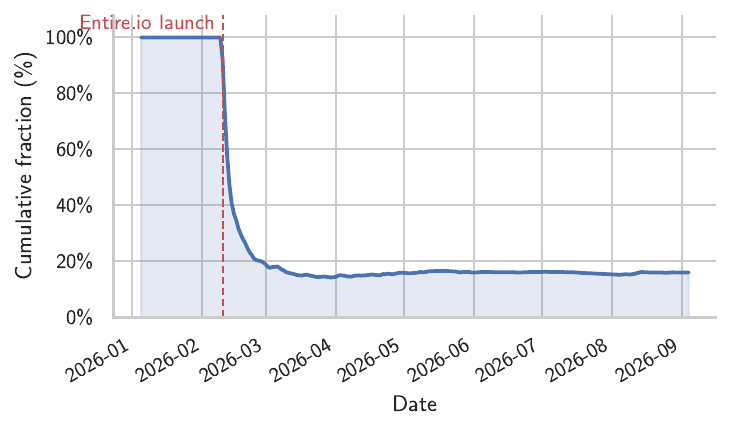}
\caption{Cumulative fraction of sessions originating from the \href{https://github.com/entireio/cli}{entireio/cli} repository over time.
Each point shows the running proportion of all sessions collected up to that date that came from this single repository.
The dashed red line marks the public launch of the Entire.io tool (February 10, 2026).}
\label{fig:entireio_fraction_over_time}
\end{figure}

Following the public launch of Entire.io on February 10, 2026, open-source developers quickly started using the tool and pushing their coding agent session data to public GitHub repositories.
Figure~\ref{fig:entireio_fraction_over_time} tracks the cumulative fraction of sessions originating from Entire.io's own \href{https://github.com/entireio/cli}{repository}. At the time of writing, this repository contributes about 16\% of all sessions in \dataset{}.

\subsection{Topic distribution}
\label{app:Topicdistribution}

To characterize the range of tasks users bring to AI coding assistants, we perform a topic analysis on all English user prompts in \dataset{}.

\subsubsection{Topic clustering methodology}
\label{app:clustering_methodology}

Starting from all English prompts, we first remove interruption signals (e.g., ``[Request interrupted by user]''), system-injected
messages (identified by XML-tag prefixes such as \texttt{<command-message>} and \texttt{<teammate-message>}), Claude skill invocations, and image references. We then stripped fenced and inline code blocks from all remaining prompts and excluded prompts whose stripped text is shorter than 30 or exceeds 1,500 characters. Finally, we deduplicate prompts on case-insensitive stripped content.

We generate sentence embeddings using the all-mpnet-base-v2 model from SentenceTransformers~\citep{reimers-gurevych-2019-sentence}.
We embed the code-stripped prompt text rather than the raw text so that embeddings reflect the user's natural language intent rather than the syntactic structure of pasted code. Before clustering, we reduce the embedding dimensionality from 768 to 20 using UMAP.

We cluster the reduced embeddings using HDBSCAN*~\citep{campello2013density,mcinnes2017hdbscan} with \texttt{min\_cluster\_size=520}, \texttt{min\_samples=5}.
This yields 20 clusters covering 47.6\% of all prompts, with cluster sizes ranging from 534 to 4,182 (median: 1,097).
The remaining 32,835 prompts (52.4\%) are classified as noise, reflecting the diversity of coding session prompts that do not form tight semantic groups.

For each cluster, we select the 100 prompts with the highest HDBSCAN* membership probability to generate a topic description, which is shown in Figure~\ref{fig:clusters}.
The descriptions are generated by gpt-5.6-sol, using the following prompt:

\begin{promptbox}
Below are 100 user prompts sampled from a single topic cluster. Each prompt was written by a software developer using an AI coding assistant.

In a <scratchpad>, do the following:
1. List the specific technologies, libraries, and project details mentioned.
2. Set all of those aside. What **general software engineering activity** unites these prompts? Think in terms of the development lifecycle: planning, prototyping, implementing features, debugging, testing, deploying, configuring infrastructure, designing APIs, refactoring, reviewing code, etc.
3. What makes this activity more specific than just "coding"? Is it about a particular layer of the stack (frontend, backend, infra, data, auth)? A particular phase (greenfield build vs. maintenance)? A particular style of work (exploratory vs. fixing vs. migrating)?

Then produce a single cluster label (4-10 words). The label must:
- Be understandable to any software engineer without knowledge of these specific projects
- Name the engineering activity and, if relevant, the stack layer or workflow phase
- Never mention specific tools, frameworks, or project names

Format:
<scratchpad>your reasoning</scratchpad>
Label: <your label>

Prompts:
{prompts}
\end{promptbox}

\subsubsection{Findings}
\begin{figure}[h!]
\centering
\includegraphics[width=\linewidth]{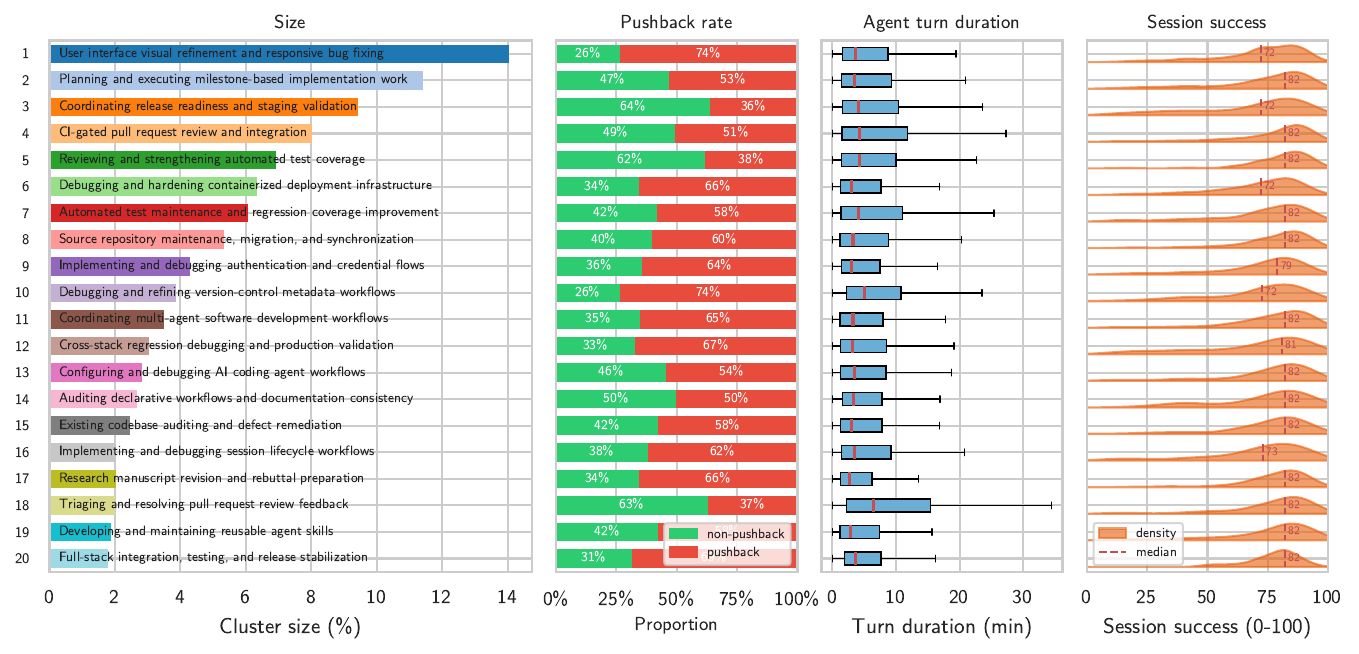}
\caption{
Topic distribution of user prompts.
Each bar represents one of the 20 clusters identified by HDBSCAN*, labeled with a GPT-generated topic summary.
The remaining panels show, per cluster, pushback rate, agent turn duration in seconds, and session success score distribution for sessions in which at least 20\% of prompts fall within the cluster.
Cluster 10 consists mostly of prompts originating from Entire.io's own \href{https://github.com/entireio/cli}{repository} (61\% of its prompts).
Some clusters (14 and 17) are dominated by a single repository and therefore may not describe very general activities.
}
\label{fig:clusters}
\end{figure}

We identify 20 topic clusters that cover 47.6\% of all prompts. The results are displayed in Figure~\ref{fig:clusters}.
Frontend UI work (cluster 1) and version-control metadata debugging (cluster 10) have the largest pushback rates (74\%).
The lowest rates are in cluster 3 and 18.

\subsection{Distribution of user personas}

\begin{figure}[h!]
\centering
    \centering
    \includegraphics[width=0.6\linewidth]{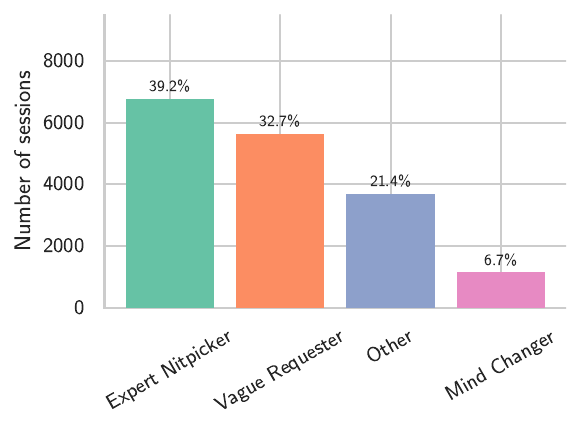}
    \caption{Distribution of user personas.}
    \label{fig:user-persona}
\end{figure}

Figure~\ref{fig:user-persona} shows the full distribution of user personas across all sessions.
Expert nitpicker is the most common persona at the session level.
The dataset contains sessions coming from 589 users.
78\% of users contribute more than one session.
When assigning each user the persona of the majority of their annotated sessions, it matches a user's individual session persona 74.5\% of the time on average.

\subsection{Coding mode distribution over time}
\label{app:coding_modes_over_time}

\begin{figure}[h!]
    \centering
    \includegraphics[width=0.8\linewidth]{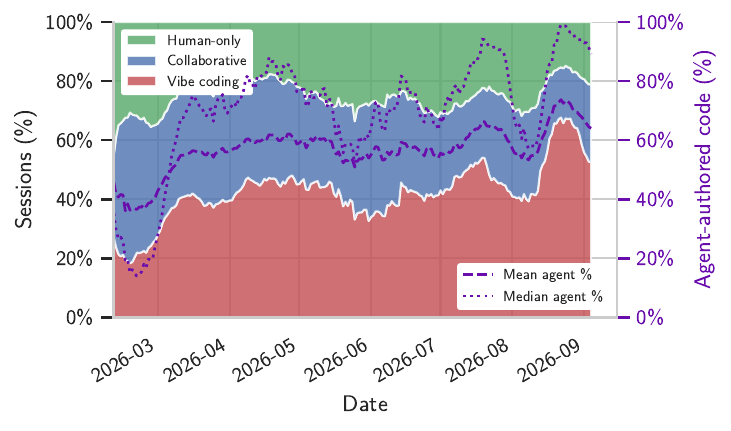}
    \caption{14-day rolling average of coding modes and agent-authored code.}
    \label{fig:coding_modes_over_time}
\end{figure}

Figure~\ref{fig:coding_modes_over_time} shows the temporal evolution of coding modes.
The share of vibe coding sessions has more than doubled since the launch of Entire's CLI tool, rising from under 20\% in February to over 50\% in recent months.

\subsection{Code vulnerability analysis with Semgrep}
\label{app:vulnerabilities}

We use Semgrep\footnote{\url{https://github.com/semgrep/semgrep}}, an open-source static analyzer that matches community-curated patterns against source code, and run it with its default \texttt{--config=auto} ruleset.
This auto-selects rules based on the languages detected in each snapshot, including Common Weakness Enumeration (CWE), which includes known types of security weaknesses~\citep{Martin2008}.
For every commit, we extract the repository state before and after the commit, scan each state, and keep only findings inside files that the commit actually modified.

\begin{figure}[h!]
    \centering
    \includegraphics[width=0.6\linewidth]{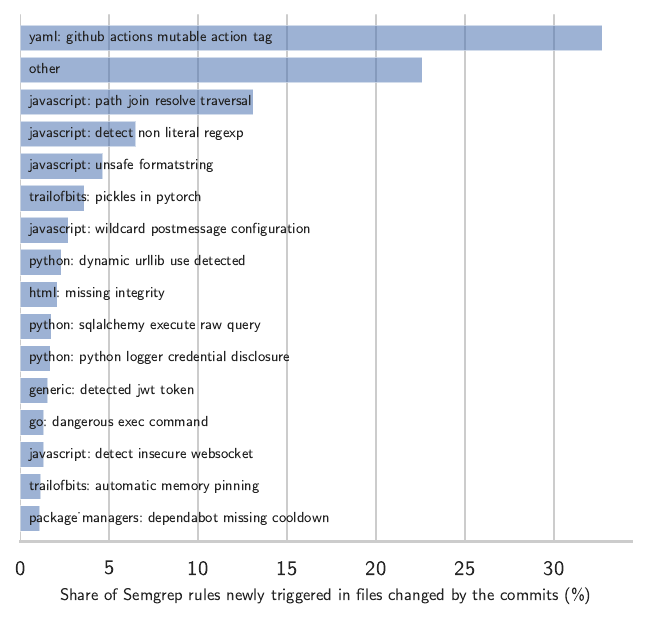}
    \caption{Distribution of introduced vulnerabilities across Semgrep rule IDs (top~15 plus \textit{other}).
   }
    \label{fig:vuln_type_distribution}
\end{figure}

\begin{figure}[h!]
    \centering
    \includegraphics[width=0.6\linewidth]{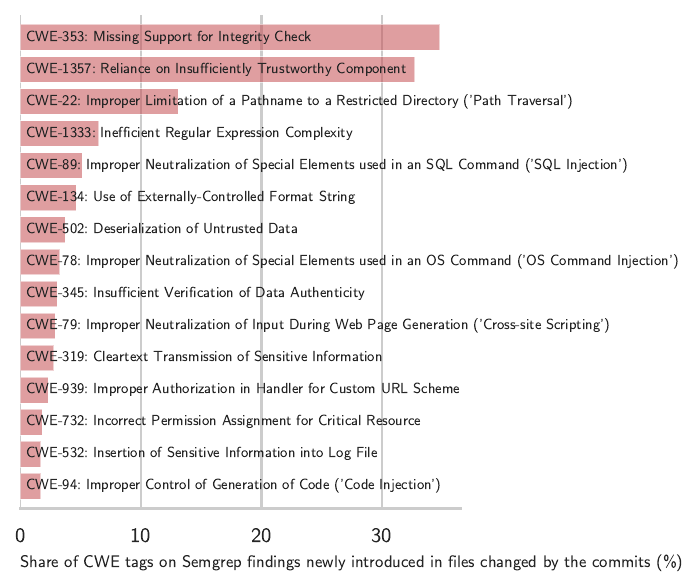}
    \caption{Distribution of introduced vulnerabilities across CWE categories (top~15 plus \textit{other}).
    }
    \label{fig:vuln_reason_distribution}
\end{figure}

\paragraph{Distribution of introduced vulnerabilities}
Figures~\ref{fig:vuln_type_distribution} and~\ref{fig:vuln_reason_distribution} break down the introduced findings by Semgrep rule and by CWE category, respectively. Two rules account for a large share of the detected vulnerabilities---mutable GitHub Actions tags (33\%) and JavaScript path joining without sanitization (13\%)---but the remaining findings include a long tail of rules and CWEs, including non-literal regular expressions (CWE-1333), externally controlled format strings (CWE-134), missing integrity checks (CWE-353), OS command injection (CWE-78), and SQL injection (CWE-89). Hence, a broad set of vulnerability types is being introduced.

\paragraph{Vulnerability example}
Figure~\ref{fig:vuln_example} shows a concrete Python example of a vulnerability introduced by a coding agent in our dataset, together with the Semgrep annotation that flags it.

\begin{figure}[h]
\centering
\begin{minipage}{0.95\linewidth}
\begin{lstlisting}[language=Python, basicstyle=\ttfamily\small,
                   frame=single, framesep=4pt, numbers=left, numbersep=6pt,
                   numberstyle=\tiny\color{gray}, breaklines=true,
                   xleftmargin=1.6em,
                   keywordstyle=\color{blue!60!black},
                   stringstyle=\color{red!60!black},
                   commentstyle=\color{gray}\itshape,
                   showstringspaces=false]
import subprocess

def run_build(target: str) -> str:
    """Run the project's build step and return stdout."""
    cmd = f"make {target}"
    # Semgrep: python.lang.security.audit.subprocess-shell-true (CWE-78)
    #   "Found subprocess function with shell=True. ...
    #    A malicious actor can inject arbitrary shell commands.
    #    Use shell=False instead."
    result = subprocess.run(
        cmd, shell=True, capture_output=True, text=True)
    return result.stdout
\end{lstlisting}
\end{minipage}
\caption{\textbf{Example of a Python vulnerability introduced by an agent in \dataset{}.} The agent builds a shell command by interpolating a user-controlled string (\texttt{target}) into an f-string and then calls \texttt{subprocess.run} with \texttt{shell=True} (line 10). The inline comment shows the Semgrep annotation, which flags a CWE-78 OS-Command-Injection risk because an attacker who can influence \texttt{target} could inject arbitrary shell commands (e.g.\ \texttt{"rm -rf \~{}"}). The standard fix is to pass arguments as a list, e.g.\ \texttt{subprocess.run(["make", target])} with \texttt{shell=False}, so they are not reparsed by a shell.}
\label{fig:vuln_example}
\end{figure}

\subsection{Agent efficiency}

\begin{figure}[h!]
\centering
\includegraphics[width=\linewidth]{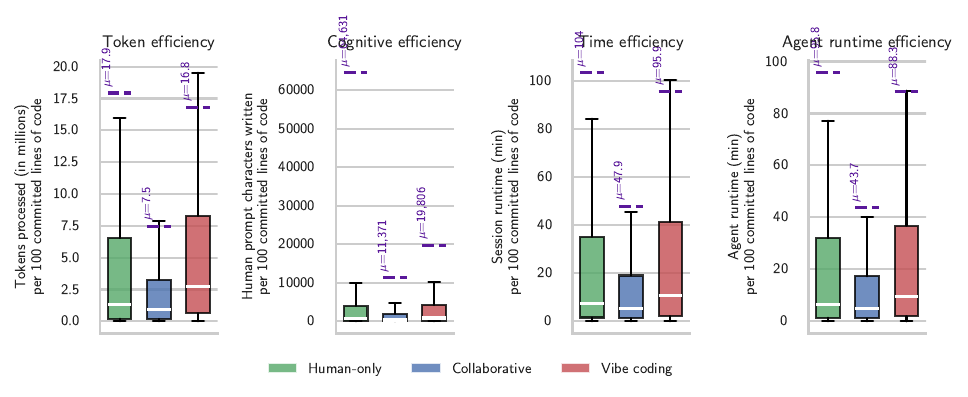}
\caption{Token, cognitive, and time efficiency per 100 committed lines of code (lower is better).
Y-axis labels describe the metrics used for the efficiency dimension.
$\mu$ indicates means.}
\label{fig:_efficiency_combined_boxplots}
\end{figure}

Figure~\ref{fig:_efficiency_combined_boxplots} compares efficiency across coding modes along four dimensions.
Vibe coding sessions are consistently less efficient: they consume roughly three times as many tokens and require more wall-clock time per 100 committed lines than collaborative sessions.
Collaborative coding achieves the best trade-off across all metrics, suggesting that human guidance helps agents produce code more economically.

\subsection{Agent turn duration over time}
\label{sec:Agent_turn_duration}

\begin{figure}[h!]
\centering
\includegraphics[width=0.85\linewidth]{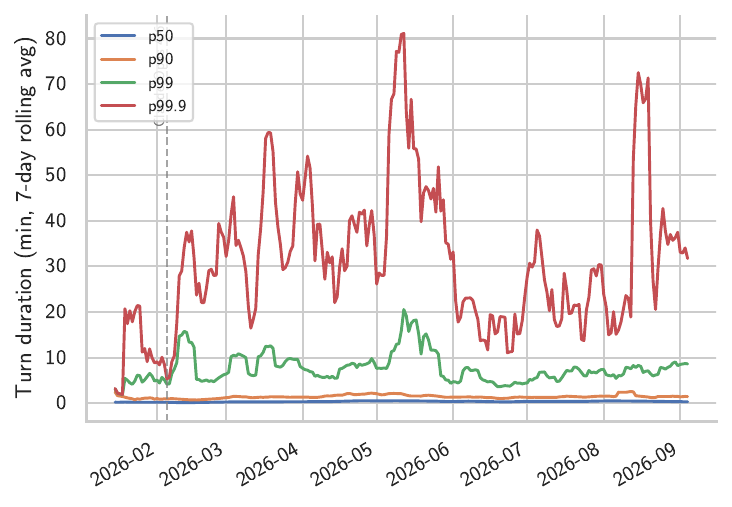}
\caption{Turn-level autonomy in Claude Code sessions. Agent turn duration in interactive Claude Code sessions. Showing 7-day rolling average of different percentiles (p50, p90, p99, p99.9) over time.}
\label{fig:turn-call-duration}
\end{figure}

Figure~\ref{fig:turn-call-duration} tracks agent turn duration over time.
Median turn durations have remained relatively stable throughout the data collection period, but the tail fluctuates. The 99.9th percentile turn lasts 55 minutes when pooling numbers over the whole period.

\subsection{Oversight rates over time}
\label{sec:oversight-over-time}

\begin{figure}[h!]
\centering
\includegraphics[width=\linewidth]{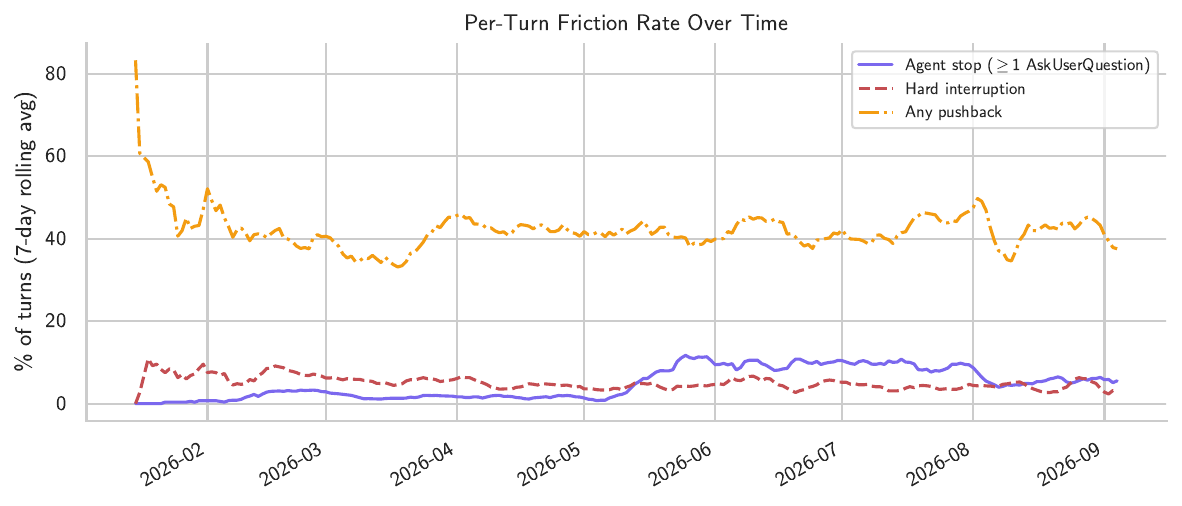}
\caption{Agent stops for clarification, user interruptions, and soft user pushback over time.}
\label{fig:clarification_rate_by_turn}
\end{figure}

Over the entire data collection period (from January to September, 2026), the shares of agent-initiated stops, user interruptions, and user pushback remain relatively stable.
We visualize this with average fractions of turns using a 7-day rolling window in Figure~\ref{fig:clarification_rate_by_turn}.

\subsection{Development activities}
\begin{figure}[thb]
\centering
\includegraphics[width=\linewidth]{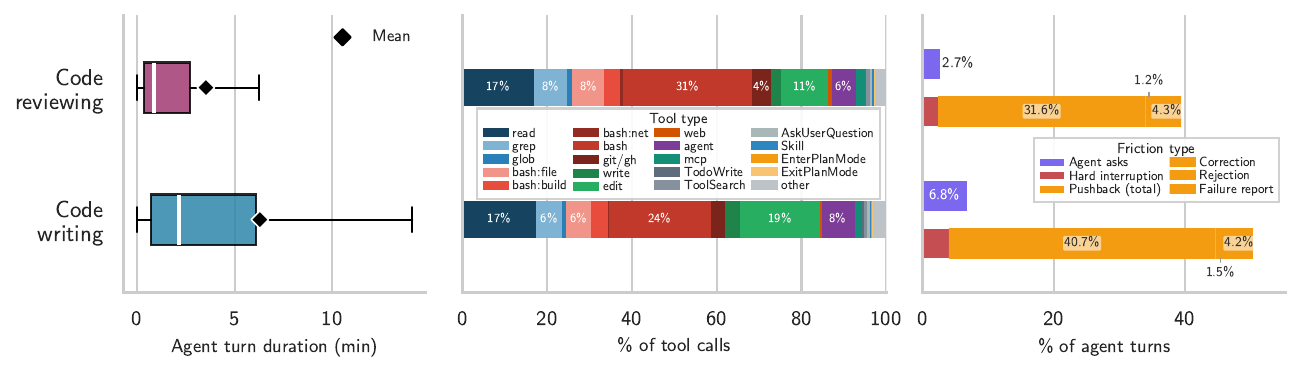}
\caption{Agent behavior by development activity (code writing vs.\ code reviewing).}
\label{fig:test_intent_unified}
\end{figure}

To examine whether agent behavior differs across development activities, we group intents into \textit{code writing} (create, refactor, connect) and \textit{code reviewing} (understand, test) prompts.
As visible in Figure~\ref{fig:test_intent_unified}, on average, \textit{code writing} prompts trigger longer agent turns (mean 6.3 vs.\ 3.5 minutes) and more file writes (3.5\% of tool calls create a new file from scratch) and edits (19\% of tool calls edit an existing file).
Furthermore, \textit{writing prompts} also elicit more friction than \textit{code reviewing} prompts: agents stop to ask questions more than twice as often (6.8\% vs.\ 2.7\% of turns), and users also interrupt and push back more frequently (see Section~\ref{sec:pushback} for more details).

\section{Data annotation}
\label{sec:annot-appendix}

Here we provide all prompts used for the final dataset annotation.
We include all validation details and prompts for the annotation tasks we crafted.
The prompt intent task is inspired by~\citet{becker2025measuring} and the user persona task is inspired by~\citet{wang2026position}.

\subsection{Validation}
\label{sec:validation}

\paragraph{Annotation codebook development and annotator agreement}

\begin{figure}[htbp]
\centering
\includegraphics[width=0.9\linewidth]{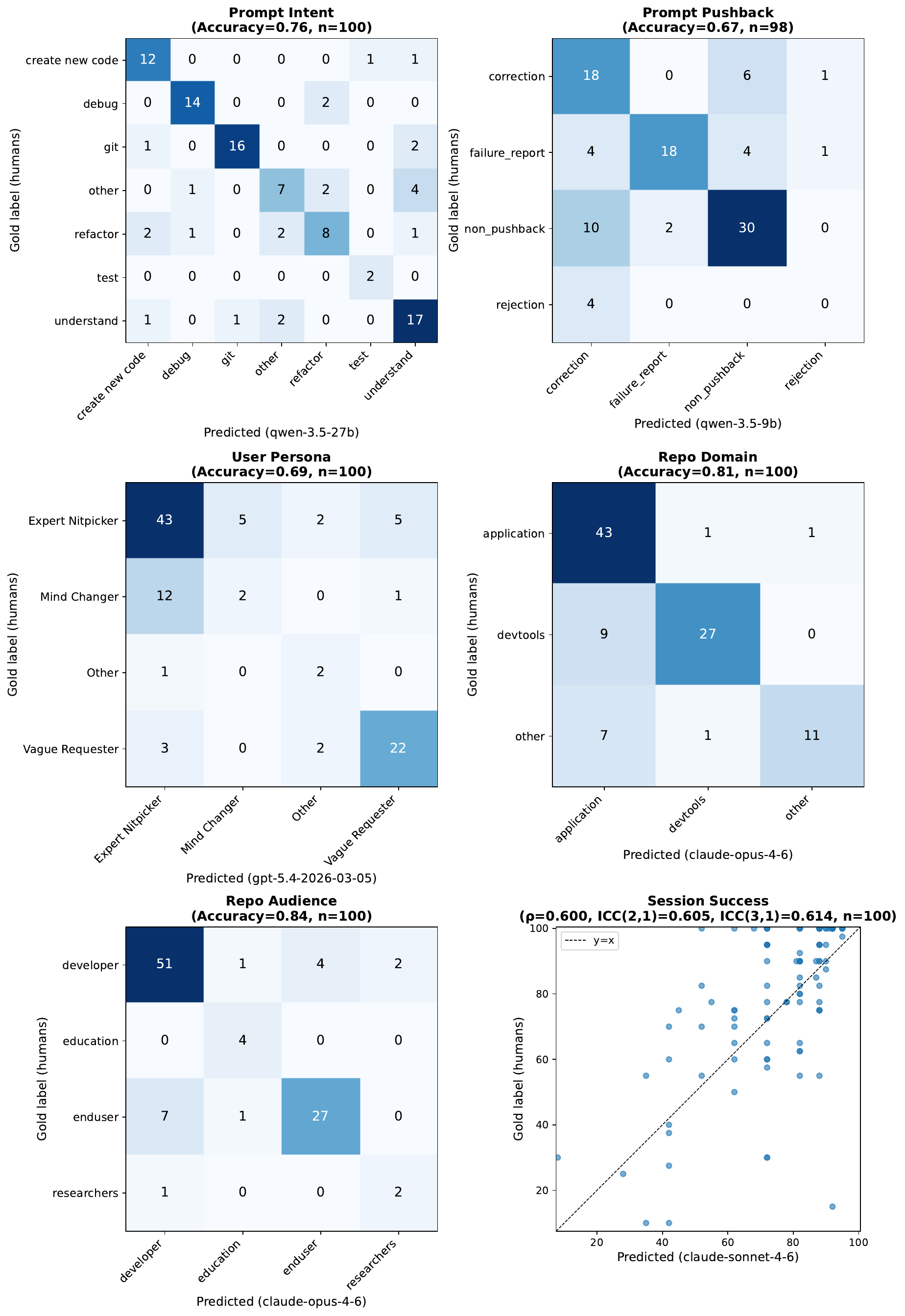}
\caption{LLM annotation agreement with gold labels from human expert annotations.}
\label{fig:combined_confusion_matrices}
\end{figure}

\begin{figure}[thb]
\centering
\includegraphics[width=0.9\linewidth]{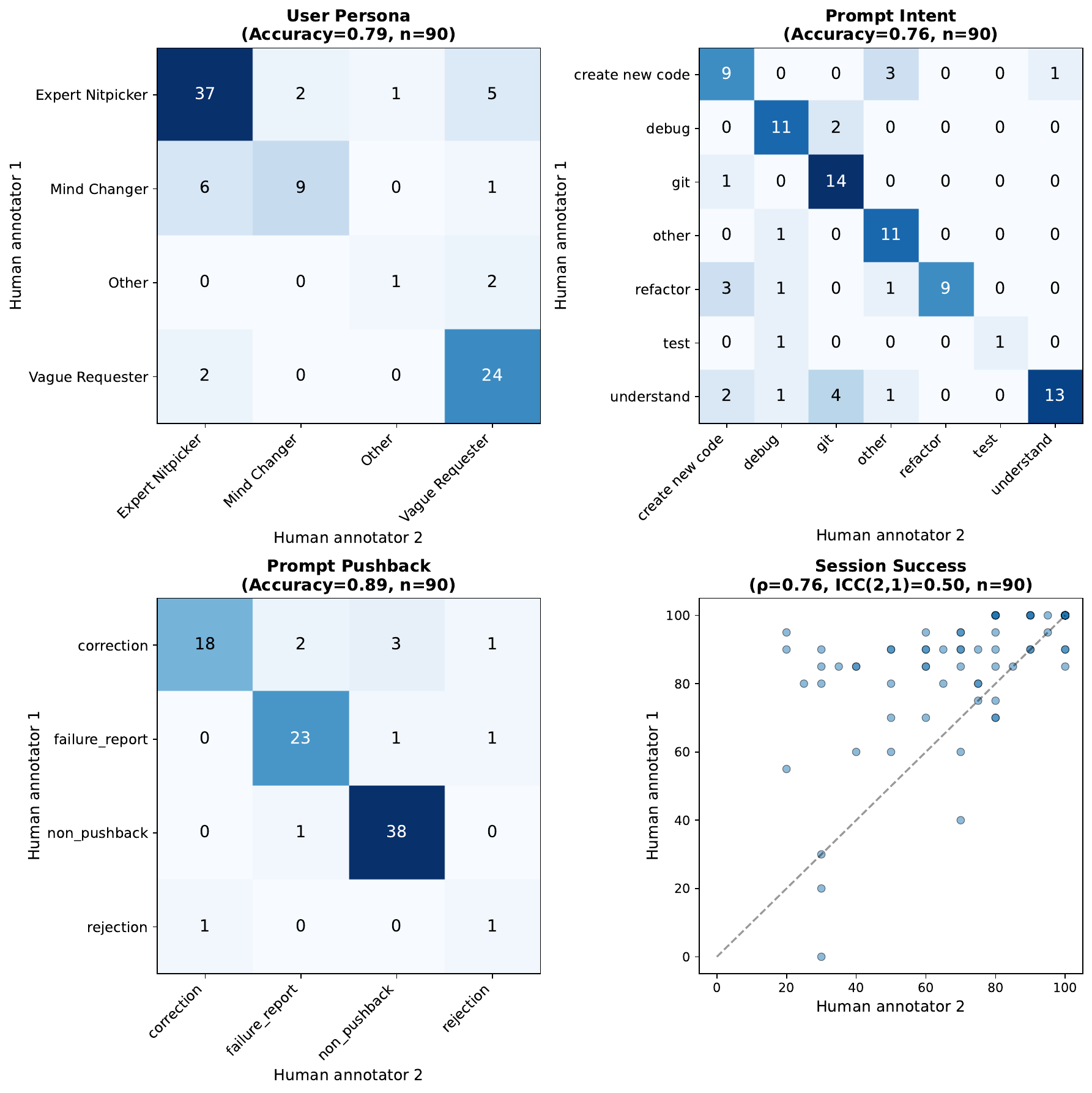}
\caption{Inter-annotator agreement confusion matrices.}
\label{fig:combined_confusion_matrices_human}
\end{figure}

To develop the annotation codebook and a dataset to test LLM annotation performance, we proceeded in three stages for each annotation task:
\begin{enumerate}
    \item First, two annotators iteratively refined the codebook until they agreed on all labels for 10 data points.
    \item Second, the same two humans proceeded to independently annotate $N_{\text{IAA}}=90$ additional data points. We computed inter-annotator agreement metrics using the results from this stage. The results in Tables~\ref{tab:iaa} and~\ref{tab:iaa_numeric} show that agreement was moderate-high for all tasks.
    This includes a binary version of the prompt pushback tasks that collapses all classification classes except the non-pushback class.
    Figure~\ref{fig:combined_confusion_matrices_human} shows the full confusion matrices for all tasks.
    For session success ratings, we discuss all cases where humans disagree by more than 20 points.
    \item Finally, the same two humans discussed all disagreements and decided on the most appropriate gold label for each data point. Together with the 10 data points from stage 1, this yielded $N_{\text{gold}}=100$ gold labels for evaluating LLM annotation performance.
\end{enumerate}
We use Cohen's $\kappa$ to measure human-human and LLM-human agreement for multi-class annotation tasks, and additionally report percentage agreement in Table~\ref{tab:iaa}~\citep{cohen1960}.
Session success is labeled with a 0--100 score, which is why we measure absolute agreement with a two-way random effect, single measurement Intraclass Correlation Coefficient, commonly referred to as ICC(2,1)~\citep{shrout1979intraclass,mcgraw1996forming}.
We additionally report Spearman $\rho$ in Table~\ref{tab:iaa_numeric} and Figures~\ref{fig:combined_confusion_matrices_human}--\ref{fig:combined_confusion_matrices}~\citep{spearman1961proof}.
We use the average of the two human annotators' session success ratings as the gold standard against which we compare the LLMs.
If human ratings differ by $>$20, the human annotators collectively decide on the most appropriate gold rating.
For the LLM vs.\ human gold rating comparison, we additionally report consistency using a two-way mixed effect, single measurement ICC, abbreviated ICC(3,1)~\citep{shrout1979intraclass}.

For the repository-level annotations, we take a slightly different approach. Namely, during stage 2, human annotator 2 reviewed the 100 repository domains and repository audience labels set by annotator 1 and either agreed with or overrode them.

All annotators are authors of this paper.

\begin{table*}[thb]
\small
\centering
\caption{Inter-annotator agreement for multi-class tasks.}
\label{tab:iaa}
\begin{tabular}{lccccc}
\toprule
\textbf{Task} & $\mathbf{N_{\text{\textbf{IAA}}}}$ & \textbf{Categories} & \textbf{Annotators} & \textbf{Agreement} & \textbf{Cohen's $\kappa$} \\
\midrule
\textbf{Prompt intent} & 90 & 7 & 2 & 68/90 (75.6\%) & 0.709 \\
\textbf{Prompt pushback} & 90 & 4 & 2 & 80/90 (88.9\%) & 0.832 \\
\textbf{Prompt pushback (binary)} & 90 & 2 & 2 & 85/90 (94.4\%) & 0.888 \\
\textbf{User persona} & 90 & 4 & 2 & 71/90 (78.9\%) & 0.662 \\
\bottomrule
\end{tabular}
\end{table*}

\begin{table*}[thb]
\small
\centering
\caption{Inter-annotator agreement for continuous session success rating task.}
\label{tab:iaa_numeric}
\begin{tabular}{lccccc}
\toprule
\textbf{Task} & $\mathbf{N_{\text{\textbf{IAA}}}}$ & \textbf{Rating scale} & \textbf{Annotators} & \textbf{Spearman $\rho$} & \textbf{ICC(2,1)}\\
\midrule
\textbf{Session success} & 90 & 0--100 & 2 & 0.757 & 0.503\\
\bottomrule
\end{tabular}
\end{table*}

\paragraph{LLM annotation performance}

We tested 9-11 LLMs and 2-4 prompt paraphrases for each task and evaluated them against the 100 human-annotated gold labels.
Table~\ref{tab:annotation_performance} shows the performance results, showing only the best-performing prompt for each model-task combination.
We then selected the model with the highest performance.
The only exception is the prompt pushback task, where we defer to the second-best-performing model, since qwen-3.5-9b offers a much better cost-performance trade-off than gpt-5.4-2026-03-05.
The high cost of this task is due to the large context: for each prompt pushback annotation, we provide not only the user message but also the full session transcript up to that point (see Table~\ref{tab:annotations} and Appendix~\ref{sec:annot-pushback}).

Figure~\ref{fig:combined_confusion_matrices} shows the full confusion matrices for all tasks.

\begin{table*}[thbp]
\small
\centering
\caption{Annotation model performance against human gold labels. For each task, we indicate the \colorbox{bestcolor}{\strut best-performing} model and the \fbox{\textbf{chosen}} model, i.e., the one we used for full-dataset annotation.
`$\circ$' indicates models that produced invalid labels and `---' denotes models that were too expensive to run.
We use accuracy (acc) for multi-class annotations and ICC(2,1) for numeric labeling tasks.
}
\label{tab:annotation_performance}
\begin{tabular}{lccccccc}
\toprule
\textbf{Model} & \rotatebox{90}{\textbf{Prompt intent (acc)}} & \rotatebox{90}{\textbf{Prompt pushback (acc)}} & \rotatebox{90}{\textbf{Pushback (binary) (acc)}} & \rotatebox{90}{\textbf{User persona (acc)}} & \rotatebox{90}{\textbf{Repository domain (acc)}} & \rotatebox{90}{\textbf{Repository audience (acc)}} & \rotatebox{90}{\textbf{Session success (ICC(2,1))}} \\
\midrule
gpt-5.4-2026-03-05 & 0.69 & \colorbox{bestcolor}{0.74} & \colorbox{bestcolor}{0.83} & \fbox{\colorbox{bestcolor}{\textbf{0.69}}} & 0.77 & 0.79 & 0.56 \\
gpt-5-mini-2025-08-07 & 0.74 & 0.64 & 0.73 & \colorbox{bestcolor}{0.69} & 0.75 & 0.79 & 0.46 \\
gpt-5-nano-2025-08-07 & 0.67 & 0.63 & 0.69 & 0.62 & 0.68 & 0.79 & 0.50 \\
claude-opus-4-6 & 0.70 & --- & --- & 0.57 & \fbox{\colorbox{bestcolor}{\textbf{0.81}}} & \fbox{\colorbox{bestcolor}{\textbf{0.84}}} & --- \\
claude-sonnet-4-6 & 0.66 & --- & --- & 0.46 & 0.78 & 0.83 & \fbox{\colorbox{bestcolor}{\textbf{0.60}}} \\
claude-haiku-4-5-20251001 & 0.69 & $\circ$ & $\circ$ & 0.60 & 0.72 & 0.79 & 0.56 \\
\midrule
gpt-oss-120b & 0.71 & 0.66 & 0.74 & 0.63 & 0.66 & 0.75 & 0.51 \\
gpt-oss-20b & 0.73 & 0.63 & 0.71 & 0.59 & 0.63 & 0.69 & 0.49 \\
qwen-3.5-27b & \fbox{\colorbox{bestcolor}{\textbf{0.76}}} & 0.66 & 0.73 & 0.61 & 0.62 & 0.76 & 0.52 \\
qwen-3.5-9b & 0.72 & \fbox{\textbf{0.67}} & \fbox{\textbf{0.79}} & 0.60 & 0.55 & 0.76 & 0.47 \\
olmo-3.1-32b & 0.59 & $\circ$ & $\circ$ & 0.51 & 0.62 & 0.69 & 0.43 \\
\midrule
$N_{\text{gold}}$ & 100 & 100 & 100 & 100 & 100 & 100 & 100 \\
\bottomrule
\end{tabular}
\end{table*}

\newpage
\subsection{Annotation prompts}
\label{sec:prompts}
We now list all LLM-based annotation tasks applied to the \dataset{} dataset.

\subsubsection{Repository type classifier}
\label{sec:annot-repo-type}

\textbf{Model:} \texttt{claude-opus-4-6}.

\begin{promptbox}
You are a classifier that categorizes software repositories based on their purpose.

You are given:
- Repository metadata: name, description, primary language, topics/tags
- A README or context excerpt (may be truncated)

## Task A - Domain tags

Assign one of the domain tags from:
- application - end-user software (web/mobile/desktop/CLI apps)
- library - reusable code packaged for others to consume
- devtools - tooling for building/testing/releasing software
- other - any other category

Instructions:
- Choose exactly one, most likely label.
- If you are unsure, pick other.

## Task B - Target audience

Assign one of target audience tags from:
- enduser
- developer
- researcher
- education (i.e., educators or students)

Instructions:
- Choose exactly one, most likely label.

## Output format

Respond in valid JSON with the following format (and no other text):

{
""domain"": <application|library|devtools|other>,
""audience"": <enduser|developer|researchers|education>,
}

Repository information:
{input}
\end{promptbox}
We aggregate library and devtools into a single category called devtools, since human annotators often disagreed about which to assign.

\subsubsection{Session persona classifier}
\label{sec:annot-session-persona}

\textbf{Model:} \texttt{gpt-5.4-2026-03-05}. Parameters: $reasoning\_effort\,=\,\texttt{low}$.

\begin{promptbox}
You are given a chronological timeline of a coding session. The timeline contains:
- User prompts
- Model responses
- AI agent coding actions (file edits, tool calls, tests run, etc.)
- Commits or diffs
- Any visible user reactions to outputs

Your task is to classify the USER (not the model) into one of the following personas:

1) Expert Nitpicker
  - Deep domain knowledge; high and consistent standards
  - Gives precise, technically specific instructions or corrections
  - Notices subtle issues and requests exact adjustments
  - Goal remains stable throughout; corrections refine execution, not direction

2) Vague Requester
  - Broad strokes only; underspecified goals
  - Missing constraints; leaves many decisions to the model
  - Does not correct or redirect in detail

3) Mind Changer
  - Shifts the overall goal or requirements mid-session
  - Contradicts or reverses earlier instructions
  - Corrections change direction, not just execution details

4) Other (use only when the session is too brief to judge, or the user's behavior clearly does not match any of the above)

Disambiguation:
- Expert Nitpicker vs Mind Changer: an Expert Nitpicker corrects HOW the model executes a stable goal; a Mind Changer revises WHAT the goal is. Repeated precise corrections within the same goal = Expert Nitpicker.
- Expert Nitpicker vs Other: prefer Expert Nitpicker over Other when the user shows specific technical demands, even if corrections are infrequent.

INSTRUCTIONS:
- Analyze the full sequence of actions in the session.
- Focus on patterns across time, not a single prompt.
- Use evidence from: constraint specificity, goal stability, revision behavior, error tolerance,
  explicit vs vague expectations.
- If multiple personas seem plausible, choose the dominant behavioral pattern.
- Do not over-index on one isolated incident.
- Reserve "Other" for sessions where no pattern is discernible.

BEFORE classifying, you MUST first complete the analysis fields below. Think carefully about each one.

OUTPUT FORMAT (JSON only):
Return exactly one JSON with the following keys, in this order:
  "distinct_goals": ["<list each distinct goal or topic the user pursued during the session>"],
  "goal_changed": true/false,
  "user_response_style": "<describe: does the user mostly give detailed technical instructions, or mostly short commands/confirmations/option-picks?>",
  "label": "<one of: Expert Nitpicker | Vague Requester | Mind Changer | Other>",
  "confidence": 0.0-1.0,
  "reason": "2-5 sentence explanation referencing specific behavioral patterns observed in the timeline."

IMPORTANT: If "distinct_goals" has more than one entry AND "goal_changed" is true, the label should be Mind Changer (unless the goals are clearly sub-tasks of one coherent objective). If "user_response_style" indicates mostly short commands, confirmations, or option-picks, the label should be Vague Requester.

Do not include anything outside the JSON.

INPUT:
<input>
\end{promptbox}

\subsubsection{Prompt intent classifier}
\label{sec:annot-prompt-intent}

\textbf{Model:} \texttt{Qwen/Qwen3.5-27B}. We use suggested decoding parameters: $temperature\,=\,0.7, top\_p\,=\,0.8, top\_k\,=\,20, presence\_penalty\,=\,1.5.$

\begin{promptbox}
You are a classifier that categorizes developer prompts based on their primary intent.

Your task is to read a user prompt and assign it exactly one of the following labels:
- create new code - The user is asking to generate new functionality, write a script, implement a feature, or build something from scratch.
- refactor - The user wants to modify, restructure, optimize, clean up, or improve existing code without fundamentally changing its purpose.
- debug - The user is diagnosing an error, investigating unexpected behavior, fixing a bug, or resolving a crash.
- understand - The user is asking for an explanation, walkthrough, summary, or clarification of existing code or behavior.
- connect - The user wants to integrate systems, connect APIs, wire components together, or build glue code between tools/services.
- git - The user is asking for help with version control, branching, merging, resolving conflicts, commits, or other git-related tasks such as fixing git-related errors.
- test - The user is asking to write tests, add validation, benchmark, evaluate performance, or verify correctness.
- other - The prompt does not clearly fall into any of the above categories.

Instructions:
- Choose exactly one label.
- Select the label that best represents the primary intent of the prompt.
- If multiple intents are present, prioritize the main action requested.
- Respond in valid JSON with the following format:
{
"label": "<one of the labels above>",
"reason": "<1-2 sentence explanation of why this label was chosen>"
}
- Do not include any text outside the JSON object.

Now classify the following prompt:
{input}
\end{promptbox}

\subsubsection{User pushback classifier}
\label{sec:annot-pushback}

\textbf{Model:} \texttt{Qwen/Qwen3.5-9B}. We use suggested decoding parameters: $temperature\,=\,0.7, top\_p\,=\,0.8, top\_k\,=\,20, presence\_penalty\,=\,1.5.$

\begin{promptbox}
You are a classifier that determines whether a user prompt in a coding agent session represents pushback against the agent's preceding action, and if so, what kind.

Pushback is any prompt where the user resists, corrects, redirects, or takes over from the agent - rather than simply continuing the workflow. Use the preceding conversation context to understand what the agent just did.

Classify the prompt into exactly one of the following categories:

- correction - The user redirects the agent by providing missing context, correcting a misunderstanding, pointing out factual errors, or changing requirements/direction/scope mid-task.
  Examples: "I said X not Y", "you changed the wrong file", "actually the API uses POST not GET", "actually, let's do X instead", "forget that approach, try Y", "on second thought, skip the tests"

- rejection - The user explicitly rejects, reverts, or refuses the agent's output without providing a specific correction.
  Examples: "undo that", "revert the last change", "no", "that's wrong", "I don't want that", "put it back the way it was"

- failure_report - The user reports that the agent's output does not work: bugs, errors, test failures, or broken behavior.
  Examples: "this still doesn't work", "it's still crashing", "same error, try again", "the tests are failing", "I get a 404 now"

- non_pushback - The prompt moves the session forward normally: a new task, building on agent output, asking a question, or routine iteration.
  Examples: "now add a login page", "good, also add unit tests", "why did you use a list here?", "change the button color to blue"

Disambiguation:
- correction vs rejection: correction provides a specific fix, missing information, or new direction; rejection just says "no" or "undo" without explaining what was wrong.
- failure_report vs rejection: failure_report = "it doesn't work" (something is broken); rejection = "I don't want that" (output is unwanted even if functional).

When uncertain:
- If the prompt contains words like "undo", "revert", "wrong", "broken", "doesn't work", "I said", "you missed", or "never mind", lean toward a pushback category.
- If the prompt reads like a standalone next step with no negative reaction, lean toward non_pushback.

Respond in valid JSON only:
{
"label": "<one of: correction, rejection, failure_report, non_pushback>",
"reason": "<1-2 sentence explanation>"
}

Preceding conversation context:
{context}

User prompt to classify:
{input}
\end{promptbox}

\subsubsection{Session success rating}
\label{sec:session_success}

\textbf{Model:} \texttt{claude-sonnet-4-6}.

\begin{promptbox}
You are scoring the overall success of an interactive coding session between a human user and an AI coding agent.

You will receive:
1. The full conversation transcript (user messages and agent responses).
2. A summary of tool calls made during the session (category counts, top tools).
3. Commit information, if any (commit messages and diff summaries).

## Evaluation procedure

First, analyze the session along five dimensions. For each, note concrete evidence from the transcript.

1. **Goal completion**: Did the agent fulfill what the user asked for? Identify every distinct user request or task. For each, judge whether it was fully resolved, partially resolved, or unresolved.

2. **Final session state**: How did the session end? A natural conclusion (user confirms satisfaction, moves on to a new topic, or signs off) is positive. An abrupt stop (user abandons mid-task, expresses frustration, or silently disengages after an unresolved error) is negative. Weigh the ending heavily - a session that goes well for many turns but ends in an unresolved failure should score substantially lower than one that resolves cleanly.

3. **Agent efficiency**: Did the agent make steady progress, or did it spin? Negative signals include: the user repeating the same instruction or correction, the agent retrying a failed approach without changing strategy, and unnecessary tool calls that do not advance the task. Positive signals include: appropriate use of research tools before acting, surfacing uncertainty and offering the user choices, and recovering from errors autonomously.

4. **Code and commit quality**: If code was produced, does it appear correct and complete based on the available evidence (test results, diff content, user reactions)? Were changes committed? Commits are a strong positive signal for task-oriented sessions but are not required for short advisory or exploratory sessions where no code change was the expected outcome.

5. **User experience**: Did the user have to fight the agent, or did the interaction flow naturally? Look for signs of satisfaction (thanks, approval, moving to the next task) and dissatisfaction (re-explaining, correcting, expressing frustration).

## Scoring rubric

After analyzing the five dimensions, assign a single integer score from 0 to 100.

- **90-100**: Every user request was fully resolved. The final session state is clean (code works, commits are sound, no dangling errors). The user's last messages indicate satisfaction or natural conclusion.
- **70-89**: The core user goal was met, but minor issues remain (e.g., an edge case unhandled, a small follow-up the user would still need to do). No major unresolved errors. The session ended on a reasonable note.
- **50-69**: Meaningful progress was made on the primary goal, but at least one significant sub-task is incomplete or the solution has a known defect. The user may have had to redirect the agent more than once on the same point.
- **30-49**: The agent produced some relevant output, but the primary user goal was not achieved. The codebase may be in a worse or broken state, or the user visibly gave up on the main task.
- **10-29**: Very little useful work was accomplished. The agent struggled repeatedly, required heavy user correction, or went down unproductive paths for most of the session.
- **0-9**: Nothing of value was produced. The session ended with no progress toward any user goal.

When a session is ambiguous (e.g., short, exploratory, no clear deliverable), score based on whether the agent's responses were the most helpful thing it could have provided given the input. Do not penalize the agent for the user's request being vague, but do penalize it if it failed to make productive use of whatever information was available.

## Output format

Respond with valid JSON only. Do not include any text outside the JSON object.

{
  "score": <integer 0-100>,
  "reason": "<2-4 sentences: what was accomplished, what issues were present, and why this score.>"
}

## Session data

{input}
\end{promptbox}

\end{document}